\documentclass[times,twocolumn,review,nopreprintline]{elsarticle}

\usepackage{medima_arxiv}
\usepackage{framed, multirow}

\usepackage{amssymb}
\usepackage{latexsym}
\usepackage{amsmath}
\usepackage{subfigure}
\usepackage{bbding}
\usepackage{makecell}
\usepackage{utfsym}
\usepackage{appendix}
\usepackage{booktabs}
\usepackage{float}
\usepackage{multirow}

\usepackage{url}
\usepackage{xcolor}

\usepackage{hyperref}

\definecolor{newcolor}{rgb}{.8,.349,.1}

\journal{Medical Image Analysis}

\begin{document}

\verso{Wang \textit{et al.}}

\begin{frontmatter}

\title{EndoBoost: a plug-and-play module for false positive suppression during computer-aided polyp detection in real-world colonoscopy (with dataset)}

\author[1,2]{Haoran Wang \fnref{fn1}}
\author[3,4]{Yan Zhu \fnref{fn1}}
\author[3,4]{Wenzheng Qin \fnref{fn1}}

\author[5]{Yizhe Zhang}

\author[3,4]{Pinghong Zhou}
\author[3,4]{Quanlin Li \corref{cor1}}
\author[1,2,4]{Shuo Wang \corref{cor1}}
\author[1,2]{Zhijian Song \corref{cor1}}

\fntext[fn1]{These authors contribute equally.}

\cortext[cor1]{Corresponding authors: shuowang@fudan.edu.cn, li.quanlin@zs-hospital.sh.cn, zjsong@fudan.edu.cn}

\address[1]{Digital Medical Research Center, School of Basic Medical Sciences, Fudan University, Shanghai 200032, China}
\address[2]{Shanghai Key Laboratory of Medical Image Computing and Computer Assisted Intervention, Shanghai 200032, China}
\address[3]{Endoscopy Center and Endoscopy Research Institute, Zhongshan Hospital, Fudan University, Shanghai 200032, China}
\address[4]{Shanghai Collaborative Innovation Center of Endoscopy, Shanghai 200032, China}
\address[5]{School of Computer Science and Engineering, Nanjing University of Science and Technology, Jiangsu 210014, China}

\received{xx xx 2022}
\finalform{xx xx 2022}
\accepted{xx xx 2022}
\availableonline{xx xx 2022}
\communicated{xx xx}

\begin{abstract}
The advance of computer-aided detection systems using deep learning opened a new scope in endoscopic image analysis. 
However, the learning-based models developed on closed datasets are susceptible to unknown anomalies in complex clinical environments. 
In particular, the high false positive rate of polyp detection remains a major challenge in clinical practice. 
In this work, we release the FPPD-13 dataset, which provides a taxonomy and real-world cases of typical false positives during computer-aided polyp detection in real-world colonoscopy. 
We further propose a post-hoc module EndoBoost, which can be plugged into generic polyp detection models to filter out false positive predictions. 
This is realized by generative learning of the polyp manifold with normalizing flows and rejecting false positives through density estimation. 
Compared to supervised classification, this anomaly detection paradigm achieves better data efficiency and robustness in open-world settings. 
Extensive experiments demonstrate a promising false positive suppression in both retrospective and prospective validation. 
In addition, the released dataset can be used to perform `stress' tests on established detection systems and encourages further research toward robust and reliable computer-aided endoscopic image analysis. The dataset and code will be publicly available at \href{http://endoboost.miccai.cloud}{\color{blue} http://endoboost.miccai.cloud}.
\end{abstract}


\end{frontmatter}


\section{Introduction}
\label{sec1}

With the advance of artificial intelligence (AI) in endoscopic image analysis \citep{wang2018development}, computer-aided detection (CADe) and diagnosis (CAD) systems are being incorporated into the clinical routine \citep{hann2021current}. 
In particular, computer-aided detection of polyps in the colon has attracted great interest due to its clinical importance for the early detection of colorectal neoplasia. 
The most commonly used quality metric in polyp detection is the adenoma detection rate (ADR), defined as the proportion of patients with at least one adenoma discovered in endoscopy~\citep{rex2015quality}. 
Several preliminary randomized controlled trials show that AI-assisted colonoscopy has achieved a significant improvement in the ADR compared with the conventional colonoscopy examination by endoscopists~\citep{repici2020efficacy,wang2019real, xu2021artificial, liu2020study}. 
Despite its promising ADR, the robustness of AI-assisted systems is still challenged by the complex environments during endoscopic procedures. 
False positives (FPs) have become a major concern in clinical practice, which occur when AI identifies a polyp, however, proved to be wrong. In other words, the AI-assisted system is too sensitive and could respond to background regions irrelevant to lesions.
Typical FPs of AI-assisted polyp detection include camera artifacts, intestinal walls with blood vessels, and other structures with a similar appearance.
The reported false positive rate (FPR) varied widely ranging between 1\% to 15\% depending on the definition and judgment methods \citep{hassan2020new,yamada2019development,urban2018deep,mori2018real,lee2020real}. 
The frequent occurrence of FPs leads to endoscopists' fatigue, distraction, and the need for refocusing. 
It costs additional time and effort to discriminate FPs from true positives (TPs). 
Sometimes FPs may even cause unnecessary endoscopic resection when the endoscopist lacks appropriate training. 
A recent survey identified the top research priorities in AI-assisted colonoscopy, where `reduce false positive rates for detection systems' ranks 3\textsuperscript{rd} among 59 future research questions~\citep{ahmad2021establishing}.

One important cause of FPs is the distribution shift between the training and test data.
Most learning-based models follow the closed-world assumption, which means the training set is complete and the test set comes from the same distribution. 
However, when the trained model is deployed in an open-world setting, it's inevitable to encounter unknown samples during training. 
With the underlying assumption violated, the model robustness is susceptible to out-of-distribution (OOD) samples.
It is well-known that the deep learning models could produce wrong predictions with high confidence in face of these samples \citep{hendrycks2016baseline, nguyen2015deep}. 
For the development of AI-assisted CADe system, most works \citep{ahmad2021establishing, wang2018development, urban2018deep} use deep neural networks like YOLO \citep{redmon2016you} and Faster R-CNN \citep{ren2015faster} for the detection of polyps. 
Although satisfactory performance is achieved on the development dataset, the model could attend to background regions not seen in the training set and generate FP prediction in real-world scenarios \citep{hsieh2021computer}.

To suppress such FPs, a straightforward solution is to improve the robustness by exposing the model to hard samples during training. 
For example, \cite{guo2020reduce} added human-verified FPs into the training set of a polyp detector and improved its robustness with active learning. 
However, such a solution requires re-training of the whole model when meets new types of FPs, which is not convenient for clinical practice.
Instead, another solution is to add a post-hoc module for the quality control of positive prediction and reject the FP ones \citep{cortes2016learning}. 
This paradigm seems more practical as the post-hoc module is agnostic to the polyp detector and can be updated independently. 
To develop such a quality control module, an intuitive way is to train a binary classifier on an appropriate dataset consisting of TPs and FPs. 
But the wide variety of FPs from clinical practice makes it difficult to curate such a dataset including all possible FPs. 
The discriminative classifier trained on the incomplete dataset suffers from the same aforementioned distribution shift problem. 
Meanwhile, the imbalanced occurrence of TPs and FPs makes the training prone to bias.

To tackle the above challenges, we suggest that anomaly detection (AD) approaches are more appropriate for the post-hoc quality control of positive detection. 
In the setting of AD, the TPs and FPs are considered as normal data and anomalies, respectively. 
The reduction of FPs can be formulated as an AD task that recognizes FPs from positive predictions. 
It is noted that AD approaches do not require anomaly samples during training, which is distinct from supervised classifiers. \cite{hendrycks2019deep} also showed that the utilization of a few available anomalies would significantly improve the performance, indicating data efficiency. 
Moreover, as the AD models focus on the learning of normal data, it is more robust to unknown anomalies. 
In this work, we explore a generic and practical solution for FP suppression in real-world AI-assisted colonoscopy. 
Firstly, we summarize a taxonomy of real-world FPs during the deployment of computer-aided polyp detection models and curate an annotated dataset including both TPs and FPs. 
Inspired by boosting algorithms \citep{schapire2003boosting}, we propose a plug-and-play module EndoBoost to augment the pre-trained CADe system in a post-hoc way. 
The manifold of TPs is learned with normalizing flows, which enables exact likelihood calculation in the feature space. 
Thus, the FPs can be rejected via thresholding the likelihoods. 
The main contributions of our work are summarized as follows: 

\begin{itemize}
  \item[$\bullet$] We release the False Positive Polyp Detection-13 (FPPD-13) dataset, which includes real-world cases of TPs and 13 classes of FPs with a comprehensive taxonomy. It is a novel addition to existing colonoscopy datasets and a valuable data source to benchmark and improve model robustness for clinical practice. 
\end{itemize}

\begin{itemize}
  \item[$\bullet$] We propose EndoBoost, a plug-and-play module for the suppression of FPs during polyp detection. 
  EndoBoost follows the formulation of anomaly detection and takes FPs as anomalies. Specifically, a normalizing flow is utilized for density estimation in the feature space and rejecting FPs during real-time inference.

\end{itemize}

\begin{itemize}
  \item[$\bullet$] We develop a learnable image encoder to obtain an informative feature space for the anomaly detection task. The image encoder and the normalizing flow are jointly optimized to learn the TP manifold while the FP samples are also exploited through outlier exposure.
\end{itemize}

\begin{itemize}
  \item[$\bullet$] Extensive experiments are performed on the real-world FPPD-13 dataset. The proposed EndoBoost module shows superior performance than other anomaly detection and classification approaches in terms of both data efficiency and robustness to unknown FP classes. The application of EndBoost is also demonstrated in real-world colonoscopy video analysis.
\end{itemize}

\section{Related Works}

We first survey existing colonoscopy datasets and provide a detailed comparison between FPPD-13 and other public datasets. Then different types of anomaly detection approaches are reviewed. Finally, the normalizing flow and its application in anomaly detection are introduced.

\subsection{Endoscopy datasets}

\begin{table*}[!t]
  \caption{
    Endoscopy datasets survey. We presented basic information about endoscopy datasets, including release time, size, and type of pathological findings and artifacts. All mentioned datasets are collected with standard endoscopy. The cross means the corresponding dataset does not contain pathological findings or artifacts. GI is an abbreviation for gastrointestinal. 
  }
  \renewcommand\arraystretch{2}
  \centering
  \scalebox{0.74}{
  \begin{tabular}{ccccccc}
  \toprule 
  \makecell[c]{Year} & \makecell[c]{Dataset} & \makecell[c]{Organs} & \makecell[c]{Pathological \\ Findings} & \makecell[c]{Artifacts} & \makecell[c]{Size} & \makecell[c]{Annotation} \\
  \midrule
  2015 & \makecell[c]{ASU-Mayo \\ polyp database \\ \citep{tajbakhsh2015automated}} & Lower GI & Polyps & \usym{2717} & 18,781 images & segmentation mask \\
  2015 & \makecell[c]{CVC-ClinicDB \\ \citep{bernal2017comparative}} & Lower GI & Polyps & \usym{2717} & 612 images & segmentation mask \\

  2017 & \makecell[c]{Kvasir \\ \citep{Pogorelov2017}} & Upper \& Lower GI & \makecell[l]{GI findings \\ with polyps}  & \usym{2717} & 8,000 images & category label \\

  2019 & \makecell[c]{HyperKvasir \\ \citep{Borgli2020}} & Upper \& Lower GI & \makecell[l]{GI findings \\ with polyps} & \usym{2717} & \makecell[l]{110,079 images \\ \& 374 videos} & \makecell[c]{category label (11,662 images) \\ segmentation mask (1,000 images)} \\

  2020 & \makecell[c]{Kvasir-SEG \\ \citep{jha2020kvasir}} & Lower GI & Polyps & \usym{2717} & 1,000 images & segmentation mask \\

  2020 & \makecell[c]{EDD2020 \\ \citep{ali2021deep}} & Upper \& Lower GI & \makecell[l]{GI findings \\ with polyps} & \usym{2717} & 356 images & \makecell[c]{bounding box \\ segmentation mask} \\
  
  2021 & \makecell[c]{PolyGen \\ \citep{ali2022assessing}} & Upper \& Lower GI & Polyps & \usym{2717} & 3,242 images & \makecell[c]{bounding box \\ segmentation mask} \\
  \midrule
  2019 & \makecell[c]{EAD2019 \\ \citep{ali2020objective}} & Upper \& Lower GI & \usym{2717} & \makecell[c]{Specularity, Saturation, Artifact, Blur, \\ Contrast, Bubble, Instruments} & 2,147 images & \makecell[c]{bounding box \\ segmentation mask (474 images)} \\

  2020 & \makecell[c]{EAD2020 \\ \citep{ali2021deep}} & Upper \& Lower GI & \usym{2717} & \makecell[c]{Specularity, Saturation, Artifact, Blur, \\ Contrast, Bubble, Instruments, Blood} & 2,531 images & \makecell[c]{bounding box \\ segmentation mask (169 images)} \\

  2021 & \makecell[c]{Kvasir Instrument \\ \citep{KvasirInstrument}} & Upper \& Lower GI & \usym{2717} & \makecell[c]{Instruments} & 590 images & segmentation mask \\
  \midrule
  2022 & FPPD-13 & Lower GI & Polyps & \makecell[c]{Endoscopy flush, Camera blur and artifacts, \\ Mucus and foreign bodies, Bubble, \\ Intestinal wall with blood vessel, \\ Inflammation, Bleeding, Stool,  \\ Postoperative wounds, Instruments, Folds, \\ Ileocecal valve, Appendix hole} & 2,600 images & \makecell[c]{category label \\ bounding box} \\ 
  \bottomrule
  \end{tabular}
  }
  \label{dataset_survey}

\end{table*}

In the past decade, the development of deep learning has significantly improved computer-aided endoscopic image analysis including lesion detection and segmentation \citep{ali2021deep}. Such progress relies on the availability of massive well-annotated data. 
To date, multiple endoscopy datasets are publicly available for academic research, including representative ones listed in Table \ref{dataset_survey}. 
Popular endoscopy datasets like Kvasir \citep{Pogorelov2017} and HyperKvasir \citep{Borgli2020} focus on the semantic analysis of endoscopic images, such as different categories of gastrointestinal (GI) findings. 
To better localize and segment the pathological findings, many datasets are released with lesion annotations. 
The segmentation mask of polyps are provided in the ASU-Mayo polyp database \citep{tajbakhsh2015automated}, CVC-ClinicDB \citep{bernal2015wm, bernal2017comparative} and Kvasir-SEG \citep{jha2020kvasir}. 
In addition, EDD2020 \citep{ali2021deep} and PolyGen \citep{ali2022assessing} provide annotations of both bounding boxes and segmentation masks. 
These datasets have made great contributions to the research community on improving the performance of CADe systems.

Recently, model robustness has brought increasing attention to endoscopic image analysis. Imaging artifacts and unexpected objects other than pathological findings could lead to erroneous predictions. 
In EAD2019 \citep{ali2020objective}, artifacts are annotated with bounding boxes, and a fraction of them are further labeled with segmentation masks. 
Further, EAD2020 \citep{ali2021deep} provides eight classes of artifacts, namely specularity, saturation, artifact, blur, contrast, bubble, instruments, and blood. 
Kvasir Instrument dataset also provides hundreds of images with the segmentation of surgical instruments which are frequently seen during colonoscopy \citep{KvasirInstrument}. 
These datasets were proposed for the purpose of artifact detection and removal before inputting into the CADe system.
For example, \cite{ali2021quality} developed an automatic framework to detect and segment different types of artifacts, providing a quality score and restoring frames with artifact corruption. However, we argue that such a paradigm has certain limitations: 
a) it is impractical to enumerate and remove all artifacts in real-time; 
b) some types of artifacts can hardly affect the polyp detection network and thus would not generate FPs; 
c) the artifact datasets (e.g., EAD2020) only represent a subset of real-world artifacts, so models trained on them may fail when encountering unknown types of artifacts. 
In this work, rather than the up-front artifact removal, we focus on the reduction of FPs from the perspective of post-hoc quality control. To curate a realistic FP dataset, we collected the erroneous predictions generated by a state-of-the-art (SOTA) polyp detector and reviewed by experienced endoscopists. Compared to the existing datasets, FPPD-13 is a novel dataset enabling the development of post-hoc FPs suppression.

\subsection{Anomaly detection}

Anomalies (a.k.a., outliers) in vision are images that deviate from some concepts of normality in low-level texture or high-level semantics. 
AD models are often trained solely on normal data (a.k.a., inliers) in an unsupervised manner, otherwise, the overwhelming difference in quantity between normal data and anomalies would cause severe class-imbalance issues for the supervised learning. 
The approaches of AD can be categorized into three types \citep{ruff2021unifying}: 

\noindent \textbf{Classification-based.} These methods aim to learn an enclosed decision boundary from normal data to discriminate anomalies. It is expected that normal data lie within while the anomalies are far from the decision boundary. 
For example, the objective of the minimum covariance determinant (MCD) is to find an ellipsoid that contains all normal data in input space \citep{rousseeuw1999fast}, and one-class support vector machine (OC-SVM) learns a hyperplane in high-dimensional space with kernel tricks \citep{manevitz2001one}.

\noindent \textbf{Reconstruction-based.} Reconstruction models are trained with normal data. It is assumed that the unknown anomalies are poorly reconstructed, so samples with high reconstruction errors are considered to be anomalies. 
The inputs of the reconstruction model are encoded to lower-dimensional vectors and then projected back to the original input space.
Typical deterministic reconstruction models are principal component analysis (PCA) \citep{shyu2003novel} with linear basis and autoencoders (AE) \citep{sakurada2014anomaly} built with nonlinear neural networks.
Besides, variational autoencoder (VAE) \citep{kingma2013auto} adopts a probabilistic framework, where the latent codes are sampled in a learned Gaussian distribution.

\noindent \textbf{Density-based.} Assuming that distributions of normal data and anomalies have a clear distinction, density-based methods model the probability distribution of normal data and estimate the density (i.e., likelihood) of a given sample. 
Ideally, the density estimator would assign higher likelihoods to normal data than anomalies, so the likelihood gap enables the detection of anomalies. 
Classical density-based methods like Kernel Density Estimation (KDE) \citep{hardle1990applied} and Gaussian Mixture Model (GMM) \citep{reynolds2009gaussian} can be easily adapted to AD, where KDE is more favored for its fewer parameters to be tuned than GMM. 
However, these two methods suffer from the curse of dimensionality, and deep probabilistic models are adopted to overcome this challenge. 
Normalizing flows \citep{papamakarios2021normalizing} stand out from many deep probabilistic models, featuring the advantage of exact likelihood calculation without approximation. 
Although VAE can also be used as a likelihood estimator, it only works well when the dimensionality is relatively small (e.g., less than five) because it optimizes a loose lower bound \citep{kingma2013auto}. 

In addition, distance-based methods like isolation forest (iForest) \citep{liu2008isolation} are also used in AD.
iForest divides the input space with decision trees, and assumes that the number of divisions for normal data is small while the number of divisions for anomalies is large.
When some anomalies are accessible during training, the unsupervised setting of AD can be extended to incorporate such anomalies. 
\cite{hendrycks2019deep} proposed an outlier exposure (OE) approach integrating an auxiliary loss for anomalies. Extensive experiments showed that OE can effectively improve the performance of unsupervised AD approaches.

\begin{figure*}
  \centering
  \includegraphics[scale=0.5]{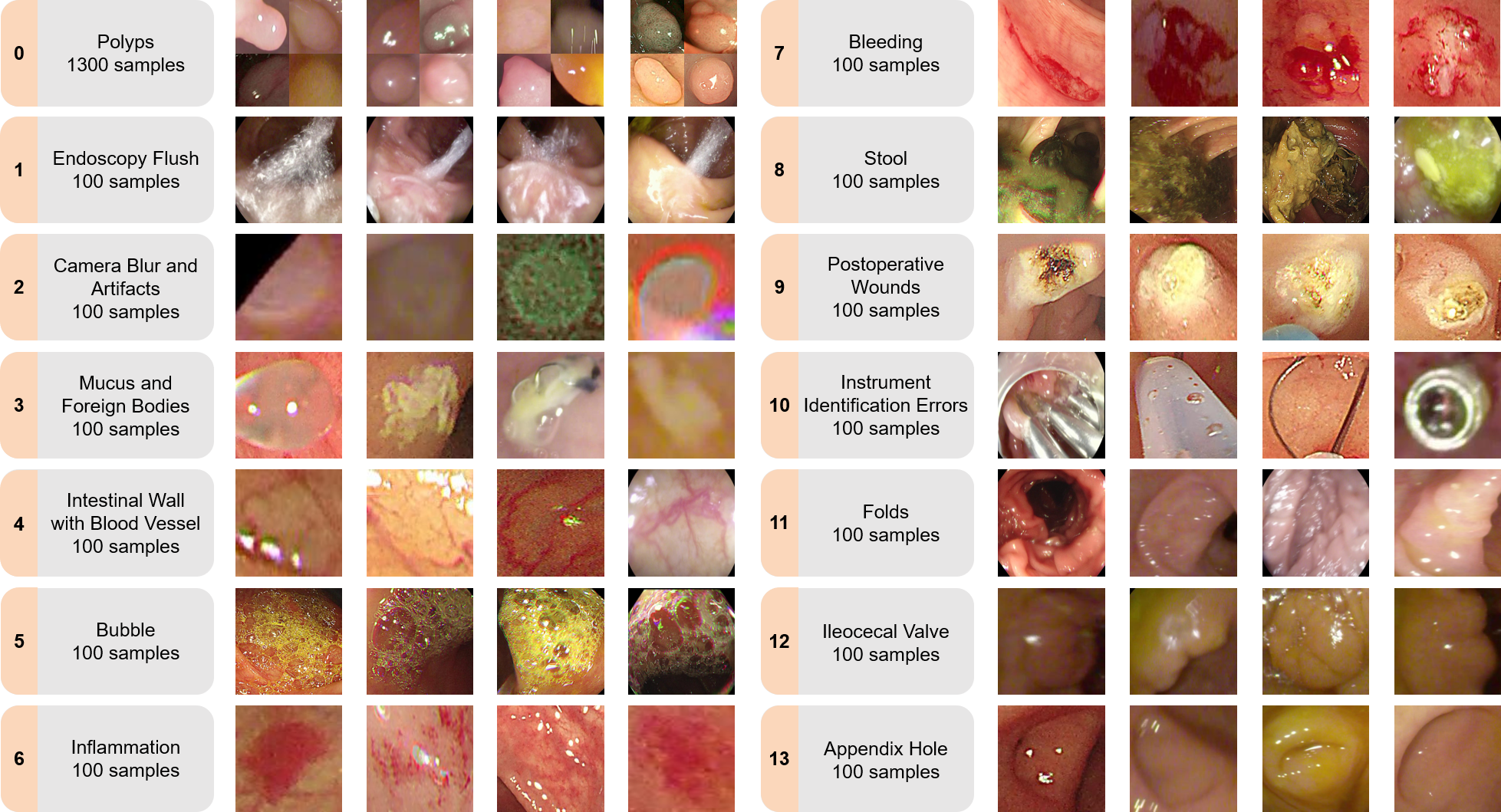}
  \caption{Taxonomy and representative cases of FPPD-13 dataset. We provide four illustrative samples for each false positive class and 16 samples for true positives. Due to the space limit, only the image content within the prediction bounding box is shown. All images are resized to square for better demonstration.}
  \label{fppd13}
\end{figure*}

\subsection{Normalizing flows}
\label{nf_rw}


Normalizing flow~\citep{kobyzev2020normalizing, papamakarios2021normalizing} is a powerful generative model to learn complex high-dimensional distributions. 
It is composed of a sequence of invertible transformation layers. Samples from the dataset can be mapped to latent codes following an analytical distribution (e.g., Gaussian distribution) and vice versa.
Also, the latent codes of normalizing flow have the same dimension as the input.
As a result, the invertible normalizing flow provides a lossless transformation between the complex data manifold and the simple analytical distribution.
With normalizing flows, data likelihood is the product of two parts: a) the likelihood of its latent code, which is easy to calculate with an analytical solution; b) the volume changes of invertible transformations evaluated by the determinant of the Jacobian matrix between input and output. 
For the training of normalizing flows, we can simply maximize the likelihood for all normal data. 

The transformation layer of normalizing flow requires invertibility and easy calculation of the determinant of Jacobian.
To meet the strict requirements above, \cite{dinh2014nice} introduced coupling layers as the basic building blocks of normalizing flows.
The coupling layer first split the input into two parts, then used additive transformation to mix these two parts and finally get the output. 
Furthermore, in their follow-up work \citep{dinh2016density}, Real-valued Non-Volume Preserving (RealNVP) extended the coupling layer with affine transformation. 
\cite{kingma2018glow} introduced 1x1 convolution as a new split strategy, achieving impressive generative quality. 
Considering the weak nonlinearity of coupling layers, \cite{behrmann2019invertible} proposed iResNet which enforced the invertibility of ResNet \citep{he2016deep} with Lipschitz constraints and provided tractable approximation to the Jacobian determinant of a residual block. 
ResFlow proposed by \cite{chen2019residual} provided a tractable unbiased density estimation on top of iResNet. 

Due to the advantages of exact density estimation, normalizing flows have been widely used in AD or other related OOD tasks \citep{rudolph2021same, cho2022unsupervised, zisselman2020deep}. 
However, \cite{nalisnick2019deep} pointed out that the normalizing flow and other deep generative models sometimes assign higher likelihoods to anomalies than normal data. 
To mitigate this issue, \cite{ren2019likelihood} used the likelihood ratio between normal data and anomalies as a score for AD. \cite{choi2018waic} introduced ensembles of generative models for a more robust likelihood estimation. 
\cite{kirichenko2020normalizing} and \cite{schirrmeister2020understanding} found that normalizing flows trained on 2D images focused on local pixel correlation, which caused overestimated likelihoods for anomalies. 
They further showed that density estimation in the one-dimensional deep semantic feature space would alleviate this issue. 
Besides, \cite{zhang2020hybrid} added a normalizing flow module to the feature extractor in the open set recognition task, achieving an improved performance of unknown class detection. These works motivate us to construct an informative feature space for the anomaly detection task of FPs.

\section{FPPD-13 Dataset}

In this section, we introduce the collection procedure of FPPD-13\footnote{This dataset is publicly available at \href{http://endoboost.miccai.cloud}{\color{blue} http://endoboost.miccai.cloud}.} and the taxonomy of real-world FPs during AI-assisted polyp detection.

\subsection{Dataset collection}

We first train a YOLOv5 \citep{yolov5} polyp detector on a private dataset collected in Zhongshan Hospital Affiliated with Fudan University, which contains endoscopic images of polyps with pathological diagnoses of colorectal hyperplastic polyps, colorectal adenomas, and colorectal cancer. Each type of pathological finding contains 5,000 images, while 10,000 images of normal colorectal mucosa backgrounds are added to the dataset. Three endoscopists with experience of more than 5,000 colonoscopies annotated the lesions with bounding boxes.
The dataset was split into training set (80\% data) and test set (20\% data) for developing and validating the YOLOv5 detector, respectively. The YOLOv5 detector achieved satisfactory performance with a sensitivity of 99.3\% and a specificity of 97.8\% in polyp detection on the held-out test set and competitive performance on the external public dataset. Details about the performance validation are provided in the Appendix. 


The well-trained YOLOv5 model was employed to analyze real-world colonoscopy videos for the collection of FPs. The colonoscopy videos started from the withdrawal once reached the ileocecum. Frames with positive predictions were captured for expert review. Two endoscopists discriminated all the positive predictions into TPs and FPs, and a third senior endoscopist was consulted for controversial images.

\subsection{Taxonomy of real-world false positives}

Referring to the previous studies and clinical experience \citep{hassan2020computer}, we divided all the FPs into 13 classes based on their different properties. Typical FP samples among the TPs and 13 FPs classes are illustrated in Fig. \ref{fppd13}. 
The FPPD-13 dataset includes a total of 2,600 representative samples. Specifically, TP samples take up half of the whole dataset, and the remaining 1,300 samples are collected evenly from the 13 FP classes, each with 100 samples. Each sample consists of an image frame of the colonoscopy video and a bounding box predicted by the YOLOv5 detector along with the class label. 
Compared to the previous EAD dataset \citep{EAD2019endoscopyDatasetI}, FPPD-13 provides more FP classes that are common during real-world colonoscopy, such as the outcome of endoscopy intervention (e.g., postoperative wounds) and anatomical background easily to be confused with polyps (e.g., ileocecal valve and appendix hole). More importantly, the samples were the ones did confuse the AI-assisted polyp detector.

\begin{figure*}[!t]
  \centering
  \includegraphics[width=0.95\textwidth]{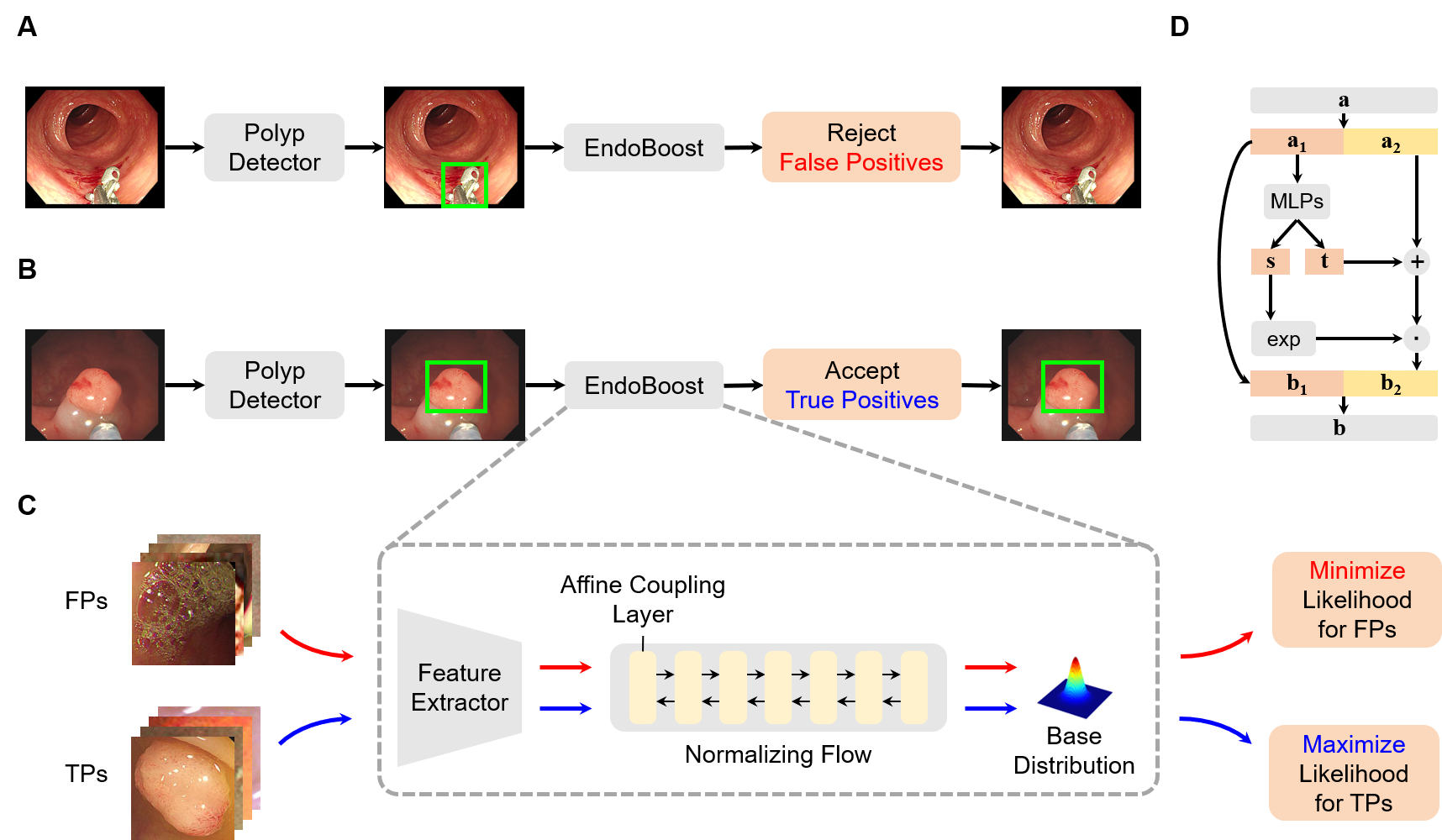}
  \caption{Schematic diagram of EndoBoost. 
  (A) Workflow of EndoBoost for False Positives. 
  (B) Workflow of EndoBoost for True Positives. 
  (C) Architecture of EndoBoost. 
  (D) Architecture of the affine coupling layer. 
  }
  \label{endoboost_overview}
\end{figure*}

\section{EndoBoost Framework}
\label{endoboost}

The workflow of the EndoBoost is shown in Fig. \ref{endoboost_overview}A\&B. 
A polyp detector is first applied to the input frame of the colonoscopy. Once the detector generates a positive prediction, the detected region within the bounding box is sent to EndoBoost for discrimination between TPs and FPs. If the EndoBoost decides that the detection is FP, the prediction bounding box will be rejected, otherwise, EndoBoost accepts it as TP.

\subsection{Problem formulation}

Given a set of positive predictions from a polyp detector $\mathcal{D} = \left\{\left(\mathbf{x}_{i}, y_{i}\right)\right\}_{i=1}^{N}$, where $\mathbf{x}_i \in \mathbb{R}^{H \times W}$ is a cropped image patch and $y \in \{0, 1\}$ is the corresponding label of TP/FP, we seek to develop a model $f:\mathbf{X} \rightarrow \mathbb{R}$ that calculates the likelihood score $s=f(\mathbf{x})$, s.t. $s_{\text{TP}}>>s_{\text{FP}}$ where $s_{\text{TP}}$ and $s_{\text{FP}}$ denote the scores for TP and FP samples, respectively.
In terms of the accessibility to FPs, there are two scenarios: a) only TPs can be used in training, i.e., $\mathcal{D} = \mathcal{D}_{\text{TP}} = \left\{\left(\mathbf{x}, y\right)|~  y=0 \right\}$; b) both TPs and FPs are available, i.e., $\mathcal{D} = \mathcal{D}_{\text{TP}} \cup \mathcal{D}_{\text{FP}}$, where $\mathcal{D}_{\text{FP}} = \left\{\left(\mathbf{x}, y\right)|~  y=1 \right\}$. 


\subsection{Network architecture}

The architecture of EndoBoost is shown in Fig. \ref{endoboost_overview}C. 
EndoBoost consists of a feature extractor $E_{\phi} = E\left(\cdot;\phi\right)$ and a normalizing-flow-based density estimation model $F_{\theta} = F\left(\cdot;\theta\right)$. 
All samples in $\mathcal{D}$ are image patches cropped with the prediction bounding box from the well-trained polyp detection model. 
The feature extractor $E_{\phi}$ maps the samples $\mathbf{x} \in \mathcal{D}$ into a $d$-dimensional feature $\mathbf{e} = E_{\phi}\left(\mathbf{x}\right) \in \mathbb{R}^d$. 
Then, the density estimation model $F_{\theta}$ transforms $\mathbf{e}$ to the latent space and estimates the likelihood $p(\mathbf{e}) = F_{\theta}\left(\mathbf{e}\right)$. 
With the above two parts combined, EndoBoost $f\left(\mathbf{x};\phi,\theta\right) = F\left(E\left(\mathbf{x};\phi\right);\theta\right)$ could discriminate the TPs and FPs by likelihood thresholding.


\subsubsection{Feature Extractor}
We adopted ResNet-50 \citep{he2016deep} as our feature extraction backbone. 
The output dimension of the feature extractor is $d$ after removing the last fully-connected (FC) layer and $d=2,048$ in this work.
The feature extractor was initialized with pre-trained weight on ImageNet \citep{deng2009imagenet}. 


\subsubsection{Density Estimation with Normalizing Flow}
\label{methods_nf}
To ensure the invertibility and fast calculation of the Jacobian determinant, normalizing flow is composed of $N$ affine coupling layers,

\begin{equation}
  \mathbf{z} = H(\mathbf{e}) = h_{N} \circ h_{N-1} \circ \cdots \circ h_{1}(\mathbf{e}),
  \label{composition}
\end{equation}

\noindent where $\mathbf{z} \in \mathbb{R}^d$ is the latent code following a Gaussian distribution, and $h_{i}$ represents the $i$-th invertible affine coupling layer, as illustrated in Fig. \ref{endoboost_overview}D. 
Each affine coupling layer splits the input into two parts and fuses them to the output. 
Given an $d$-dimensional input $\mathbf{a} \in \mathbb{R}^d$ and output $\mathbf{b} \in \mathbb{R}^d$, the affine coupling layer simply divides the input $\mathbf{a}$ in half, that is $\mathbf{a} = \left[\mathbf{a}_1, \mathbf{a}_2\right]$, where $\mathbf{a}_1 = \mathbf{a}_{1:m} \in \mathbb{R}^m$, $\mathbf{a}_2 = \mathbf{a}_{m+1:d} \in \mathbb{R}^m$ and $m = {\frac d2}$. The mapping between the input $\mathbf{a}$ and output $\mathbf{b}$ is 

\begin{equation}
  \begin{aligned}
    \mathbf{b}_{1} &= \mathbf{a}_{1} \\ 
    \mathbf{b}_{2} &= \exp\left(\mathbf{s}\right)\cdot \left[ \mathbf{a}_{2} + \mathbf{t} \right]
  \end{aligned}
\end{equation}

\noindent where $\mathbf{b} = \left[\mathbf{b}_1, \mathbf{b}_2\right]$, $\mathbf{b}_1 = \mathbf{b}_{1:m}$ and $\mathbf{b}_2 = \mathbf{b}_{m+1:d}$, $\mathbf{s} = g_s\left(\mathbf{a}_{1}\right) \in \mathbb{R}^m$ and $\mathbf{t} = g_t\left(\mathbf{a}_{1}\right) \in \mathbb{R}^m$, $g_s$ and $g_t$ are both multi-layer perceptrons (MLPs) with $L$ layers. 
The inverse mapping of such affine coupling layer is analytical: 

\begin{equation}
  \begin{aligned}
    \mathbf{a}_{1} &= \mathbf{b}_{1} \\ 
    \mathbf{a}_{2} &= \exp\left(-\mathbf{s}\right)\cdot \mathbf{b}_{2} - \mathbf{t}
  \end{aligned}
\end{equation}
The Jacobian matrix of an affine coupling layer is 

\begin{equation}
  \frac{\partial \mathbf{b}}{\partial \mathbf{a}} = \left[\begin{array}{ccc}
  \frac{\partial \mathbf{b}_{1}}{\partial \mathbf{a}_{1}} &  \frac{\partial \mathbf{b}_{1}}{\partial \mathbf{a}_{2}} \\
  \frac{\partial \mathbf{b}_{2}}{\partial \mathbf{a}_{1}} & \frac{\partial \mathbf{b}_{2}}{\partial \mathbf{a}_{2}}
  \end{array}\right]= \left[\begin{array}{ccc}
  \mathbf{I} &  \mathbf{0} \\
  \frac{\partial \mathbf{b}_{2}}{\partial \mathbf{a}_{1}} & \text{diag}\left(\exp\left(\mathbf{s}\right)\right)
  \end{array}\right],
\end{equation}

\noindent where $\mathbf{I}$ is the identity matrix and $\text{diag}\left(\cdot\right)$ is the diagonal matrix. Since the determinant of a lower triangular matrix is the product of its diagonal elements, the Jacobian determinant of an affine coupling layer is 

\begin{equation}
\log\left\vert\det\frac{\partial \mathbf{b}}{\partial \mathbf{a}}\right\vert = \sum_j s_j
\end{equation}

Let denote the distribution of TPs in the feature space as $p\left(\mathbf{e}\right)$. The normalizing flow model provides a convenient way to calculate the log-likelihood with the change-of-variables formula:
\begin{equation}
  \log p(\mathbf{e}) = \log p(\mathbf{z})+\log\left|\operatorname{det} \frac{\partial \mathbf{z}}{\partial \mathbf{e}}\right|,
  \label{logq}
\end{equation}
\noindent where $p(\mathbf{z}) = \mathcal{N}(\mathbf{z};\textbf{0}, \textbf{I})$, $\left\vert\operatorname{det}\right\vert$ is the absolute value of determinant, and $\frac{\partial \mathbf{z}}{\partial \mathbf{x}}$ is the Jacobian matrix between the input and output of normalizing flow. Let denote $\mathbf{h}_0=\mathbf{e}$, $\mathbf{h}_n=h_{n} \circ \cdots \circ h_{1}(\mathbf{e})$ and $\mathbf{h}_N=\mathbf{z}$, Jacobian matrix of the composed transformation can be derived according to the chain rule, 
\begin{equation}
  \frac{\partial \mathbf{z}}{\partial \mathbf{e}}=\frac{\partial \mathbf{h}_N}{\partial \mathbf{h}_0}=\frac{\partial \mathbf{h}_N}{\partial \mathbf{h}_{N-1}}\frac{\partial \mathbf{h}_{N-1}}{\partial \mathbf{h}_{N-2}}\cdots\frac{\partial \mathbf{h}_1}{\partial \mathbf{h}_0}
\end{equation}
\noindent and the absolute value of determinant of $\frac{\partial \mathbf{z}}{\partial \mathbf{e}}$ is 

\begin{equation}
  \left\vert\text{det} \frac{\partial \mathbf{z}}{\partial \mathbf{e}}\right\vert=\left\vert\text{det} \frac{\partial \mathbf{h}_N}{\partial \mathbf{h}_{N-1}}\right\vert \cdot \left\vert\text{det} \frac{\partial \mathbf{h}_{N-1}}{\partial \mathbf{h}_{N-2}}\right\vert \cdots\left\vert\text{det} \frac{\partial \mathbf{h}_1}{\partial \mathbf{h}_0}\right\vert=\prod_{i=1}^N \left\vert\text{det} \frac{\partial \mathbf{h}_i}{\partial \mathbf{h}_{i-1}}\right\vert
\end{equation}

\noindent Eq. \ref{logq} of multiple composed transformation can be rewritten as 

\begin{equation}
  \log p(\mathbf{e})=\log p\left(\mathbf{z}\right)+\sum_{i=1}^N \log \left|\text{det} \frac{\partial \mathbf{h}_i}{\partial \mathbf{h}_{i-1}}\right|
\end{equation}

\subsection{Loss functions}
With the normalizing flow acting as a density estimator, EndoBoost calculates the likelihood score of each input sample $\mathbf{x}$. The network is trained with maximizing likelihoods for TPs and minimizing likelihoods for FPs. 

\noindent \textbf{Maximum Likelihood Estimation (MLE) for TPs.} 
We assume that the TPs and FPs follow different distributions in the $d$-dimensional feature space.
MLE is used to optimize the parameter $\theta$ of normalizing flow, which maximizes the expectation of the log-likelihood of all the given TPs observations from $\mathcal{D}_{\text{TP}}$, that is 

\begin{equation}
  \underset{\theta}{\max} \mathbb{E}_{\mathbf{x} \sim \mathcal{D}_{\text{TP}}} \log p(\mathbf{e})
\end{equation}

\noindent The loss function for TPs is

\begin{equation}
  \mathcal{L}_{\text{TP}}\left(\theta, \phi\right) = - \mathbb{E}_{\mathbf{x} \sim \mathcal{D}_{\text{TP}}} \log p(\mathbf{e})
  \label{loss_tp}
\end{equation}

\noindent \textbf{Minimizing Likelihoods for FPs.} We expect the log-likelihoods of the TPs to be higher than FPs. When some FPs are available, i.e., $\mathcal{D} = \mathcal{D}_{\text{TP}} \cup \mathcal{D}_{\text{FP}}$, the likelihood gap can be widened through outlier exposure. Given the FPs subset $\mathcal{D}_{\text{FP}}$, a margin loss is adopted to minimize the likelihood for FPs 
\begin{equation}
  \mathcal{L}_{\text{FP}}\left(\theta, \phi\right) = \mathbb{E}_{\mathbf{x} \sim \mathcal{D}_{\text{FP}}} \max\left(0, \log p(\mathbf{e}) - \epsilon\right)
  \label{loss_fp}
\end{equation}
where $\epsilon$ is the margin parameter that controls the likelihood gap between TPs and FPs.

\noindent \textbf{Joint optimization with feature extractor.} Previous studies suggested that an informative feature space benefits the anomaly detection task. In this work, we jointly optimize feature extractor $E_{\phi}$ as well as the normalizing-flow-based $F_{\theta}$ in an end-to-end manner. With Eq.\ref{loss_tp}\&\ref{loss_fp}, the total loss to optimize EndoBoost is 
\begin{equation}
\begin{split}
   \mathcal{L}\left(\theta, \phi\right) &= \mathcal{L}_{\text{TP}}\left(\theta, \phi\right) + \mathcal{L}_{\text{FP}}\left(\theta, \phi\right) \\
      &= - \mathbb{E}_{\mathbf{x} \sim \mathcal{D}_{\text{TP}}} F_{\theta}\left(E_{\phi}\left(\mathbf{x}\right)\right) + \mathbb{E}_{\mathbf{x} \sim \mathcal{D}_{\text{FP}}} \max\left(0, F_{\theta}\left(E_{\phi}\left(\mathbf{x}\right)\right) - \epsilon\right)
\end{split}
   \label{endoboost_final_loss}
\end{equation}

\subsection{Variants of EndoBoost}
Depending on the accessibility of FPs and whether to optimize the feature extractor, EndoBoost has three variants:

\begin{itemize}
  \item \textbf{EndoBoost-MLE.} Only TPs are used for the network training, and the parameters of the pre-trained feature extractor are frozen. The training loss is $\mathcal{L}\left(\theta\right) = \mathcal{L}_{\text{TP}}\left(\theta\right)$.  
  \item \textbf{EndoBoost-Frozen.} Both TPs and FPs are used during training, while the feature extractor is pre-trained and fixed, thus the training loss is $\mathcal{L}\left(\theta\right) = \mathcal{L}_{\text{TP}}\left(\theta\right)+ \mathcal{L}_{\text{FP}}\left(\theta\right)$.  
  \item \textbf{EndoBoost-Finetune.} This variant utilizes both TPs and FPs for the end-to-end joint training of the feature extractor and the normalizing flow model. This makes the most use of the FPPD-13 dataset and the training loss is Eq.\ref{endoboost_final_loss}.
\end{itemize}

\section{Experimental design}

\subsection{Dataset}
EndoBoost variants were validated and compared to other competitors on the proposed FPPD-13 dataset. We adopted five-fold cross-validation to reduce the randomness caused by data split. In each fold, a fraction of the training set is randomly sampled for internal validation, so that the ratio between training, validation and test set is 7:1:2. For any evaluation metric, we report the mean and standard deviation of all five folds.

\subsection{Experimental setup}

To demonstrate the use of FPPD-13 dataset and validate EndoBoost under different clinical scenarios, we performed three experiments as follows:

\noindent \textbf{Comparative experiments.} The purpose of comparative experiments was to benchmark different AD methods when only normal data are available. In other words, only samples of TPs in the training set could be used. 

\noindent \textbf{Data-efficiency experiments.} This setup aimed to explore the data efficiency when FPs are accessible during training. A data-efficient method is expected to use as few FPs as possible to achieve the highest possible performance in the test set. All TPs and a portion of randomly sampled FPs were available for training, with the sampling ratios of FPs being 1\%, 5\%, 10\%, 20\%, 30\%, 40\%, 50\%, 60\%, 70\%, 80\%, 90\%, and 100\%. 

\noindent \textbf{Class-robustness experiments.} In this experiment, we were interested in the model robustness to unknown classes of FPs. Given an FPs class $c$, all TPs and the rest 12 classes were accessible for training, while the validation set and test set included all TPs and only FPs from class $c$. 

For the above experiments, performance on the test set was reported using the model with the best performance on the validation set. Note that, the amounts of TPs and FPs in the test set were balanced in the comparative and data-efficiency experiments, however, imbalanced in class-robustness experiments. 

\subsection{Comparison Methods}

In comparative experiments, we used EndoBoost-MLE variant since only TPs were available for training. Competitors cover most categories of AD methods, including KDE, OC-SVM, PCA, AE, VAE, and iForest. For a fair comparison, the same feature extractor pre-trained on ImageNet was used to generate input features for all methods.

In the data-efficiency and class-robustness experiments, we compared different types of post-hoc approaches. EndoBoost represented the anomaly detection approach. Since FPs were partly accessible during training, we used a ResNet classifier as the competitor representing the binary classification. For a fair comparison, EndoBoost used the same ResNet backbone architecture for feature extraction. To explore how much the feature space affects the performance, we also compared three variants for ResNet classification:

\begin{itemize}
  \item \textbf{ResNet-Frozen-SVM.} An SVM classifier was trained to distinguish TPs and FPs based on the ImageNet pre-trained features.
  \item \textbf{ResNet-Frozen-FC.} Except for the last layer, all the other layers of ResNet were frozen. The last FC layer was trained to discriminate TPs and FPs.
  \item \textbf{ResNet-Finetune.} All weights of ResNet were finetuned from the initialization of ImageNet pre-trained weight.
\end{itemize}

\subsection{Evaluation metrics}

Because of the class-imbalance issue between TPs and FPs, appropriate metrics are needed to evaluate the performance of EndoBoost and all the competitors. Average precision (AP) and area under the receiver operating characteristic curve (AUC) are threshold-independent and served as the main indicators of FP suppression performance in this work. We also used threshold-dependent metrics like accuracy, precision, sensitivity (recall), and specificity to evaluate the performance of different methods. To determine a proper threshold, we calculated the F1 score of every point on the precision-recall (PR) curve, and the threshold with the highest F1 score was used.

\subsection{Implementation details}

We cropped the image content within the predicted bounding box and resized it to the shape of 224x224, to adapt the ResNet-50 backbone. 
All models were implemented with PyTorch \citep{paszke2019pytorch} and Scikit-learn \citep{scikit-learn}, on a workstation with an NVIDIA GeForce RTX 3090 (24GB RAM) GPU and an Intel(R) Core(TM) i9-12900K CPU. 

For the network architecture, the normalizing flow part of EndoBoost consists of $N=32$ affine coupling layers, $g_s$ and $g_t$ in each affine coupling layer contain one FC layer with the hidden dimension of 512. All variants of EndoBoost were trained with AdamW \citep{loshchilov2017decoupled} in a learning rate of 1e-5 and weight decay of 1e-1 for 100 epochs. The batch size of EndoBoost-MLE and EndoBoost-Frozen was 2,048 while the batch size of EndoBoost-Finetune was 32 because updating the weights of the feature extractor and density estimator simultaneously requires more GPU memory. For ResNet binary classifier, the ResNet-50 backbone was used and the output dimension of the last FC layer was reduced to two for ResNet-Frozen-Linear and ResNet-Finetune. All variants of ResNet were optimized with AdamW with a learning rate of 1e-2 and weight decay of 1e-3 for 100 epochs. The batch size of ResNet binary classifiers was 128.

\section{Results}
\subsection{Benchmark of different anomaly detection methods}

\begin{table*}[!t]
  \caption{
    Quantitative results of comparative experiments on the FPPD-13 dataset, in which only the TPs are available. Numbers in parentheses of the reconstruction-based methods are the reduced dimensionality. The best results of each metric are shown in bold, second-best results are underlined in italics. 
  }
  \centering
  \scalebox{0.8}{
  \begin{tabular}{ccccccc}
  \toprule
  & AP $\uparrow$ & AUC $\uparrow$ & Accuracy $\uparrow$ & Precision $\uparrow$ & Sensitivity $\uparrow$ & Specificity $\uparrow$   \\ \midrule
  EndoBoost-MLE & \textbf{0.788$\pm$0.017} & \textbf{0.793$\pm$0.020} & \textbf{0.705$\pm$0.019} & \textbf{0.655$\pm$0.026} & 0.877$\pm$0.061 & \textbf{0.534$\pm$0.078} \\
  KDE       & 0.646$\pm$0.024 & 0.610$\pm$0.027 & 0.569$\pm$0.021 & 0.541$\pm$0.016 & 0.928$\pm$0.036 & 0.209$\pm$0.076 \\
  OC-SVM     & \textit{\underline{0.733$\pm$0.011}} & 0.726$\pm$0.004 & 0.627$\pm$0.037 & 0.588$\pm$0.036 & 0.886$\pm$0.071 & 0.368$\pm$0.145 \\
  MCD       & 0.700$\pm$0.009 & \textit{\underline{0.730$\pm$0.004}} & 0.657$\pm$0.009 & 0.604$\pm$0.011 & 0.915$\pm$0.036 & 0.398$\pm$0.051 \\
  PCA (10)  & 0.654$\pm$0.020 & 0.677$\pm$0.013 & 0.611$\pm$0.017 & 0.567$\pm$0.012 & 0.948$\pm$0.024 & 0.275$\pm$0.047 \\
  PCA (100) & 0.690$\pm$0.013 & 0.721$\pm$0.005 & \textit{\underline{0.657$\pm$0.008}} & \textit{\underline{0.605$\pm$0.008}} & 0.902$\pm$0.030 & \textit{\underline{0.412$\pm$0.035}} \\
  AE (128) & 0.647$\pm$0.031 & 0.675$\pm$0.025 & 0.605$\pm$0.022 & 0.561$\pm$0.014 & \textbf{0.960$\pm$0.013} & 0.249$\pm$0.044 \\
  VAE (128) & 0.718$\pm$0.008 & 0.703$\pm$0.008 & 0.588$\pm$0.028 & 0.557$\pm$0.025 & 0.902$\pm$0.062 & 0.275$\pm$0.118 \\
  iForest   & 0.644$\pm$0.034 & 0.634$\pm$0.026 & 0.577$\pm$0.015 & 0.544$\pm$0.010 & \textit{\underline{0.953$\pm$0.025}} & 0.200$\pm$0.046 \\ \bottomrule
  \end{tabular}
  }
  \label{comp_tab}
\end{table*}

We first evaluated the performance of different AD models when only TPs were accessible during training.  As shown in Table \ref{comp_tab}, EndoBoost-MLE achieved the highest AP (0.788) and AUC (0.793) among all the AD models. Interestingly, the lowest sensitivity (0.877) but highest precision (0.655) implicates a conservative behavior of EndoBoost-MLE in rejecting positive predictions. It should be noted that precision is more important than sensitivity due to the adverse outcome of missing polyps. In other words, models with better confidence (i.e., precision) in rejecting FPs are preferred. However, the overall low precision of these AD models indicates that the outlier exposure during training is necessary to develop a practical quality control module, which is one motivation of the FPPD-13 dataset.

\subsection{Data-efficiency of outlier exposure}

\begin{figure*}[!ht]
  \centering
  \includegraphics[width=\textwidth]{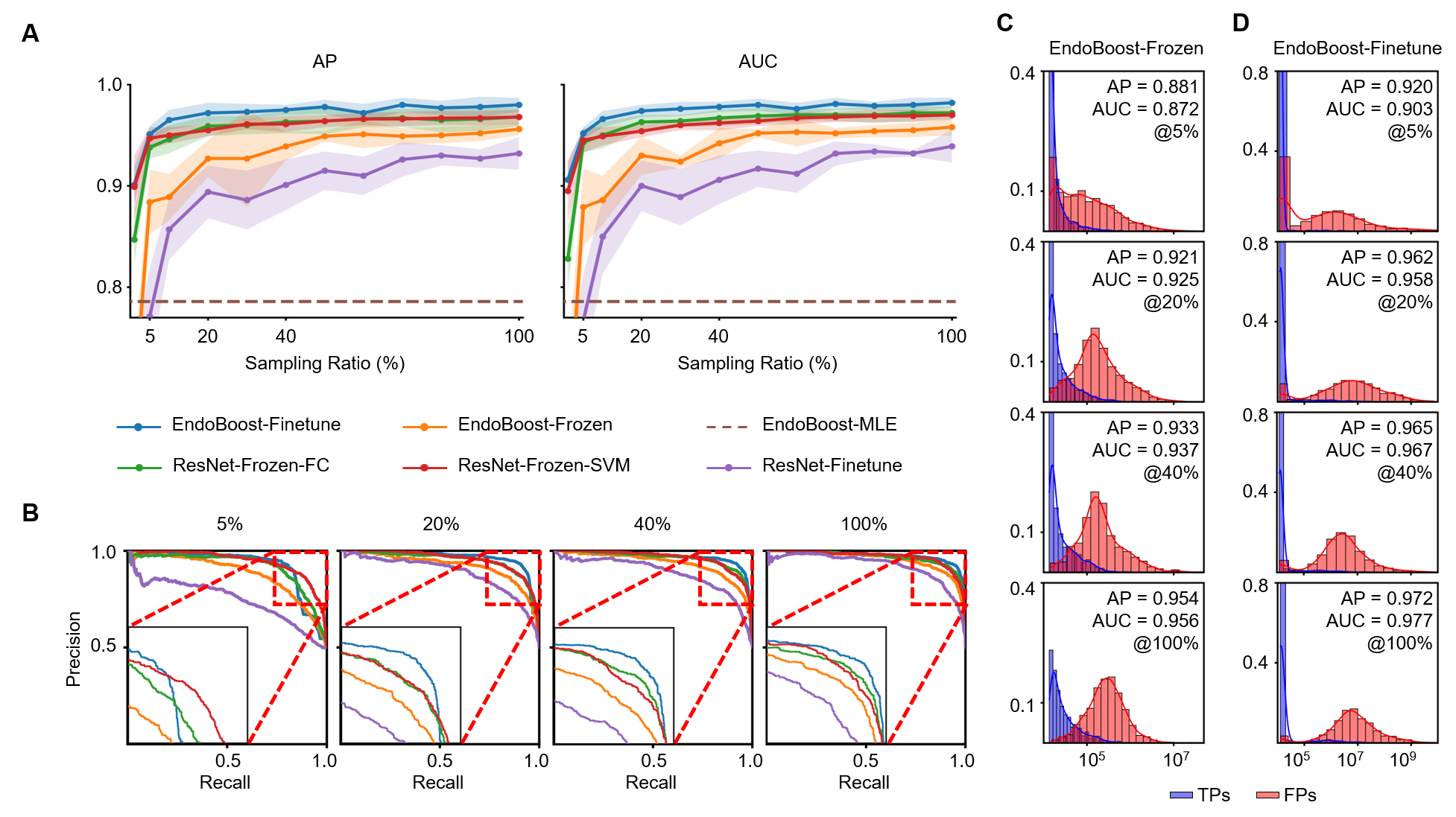}
  \caption{
    Quantitative results for data-efficiency experiments.
    (A) AP and AUC of data-efficiency experiments at all sampling ratios. All methods are shown in different colors. The average of all cross-validation folds is reported and the shaded area reflects standard error. The performance of EndoBoost-MLE is shown in dashed since it cannot utilize FPs for training. 
    (B) PR curves of data-efficiency experiments. PR curves present a more detailed comparison between EndoBoost and other competitors. 
    (C) NLL histograms of EndoBoost-Frozen at selected sampling ratios. The TPs are shown in blue while the FPs are shown in red. The x-axis is in log-scale and KDE curves are plotted for smoothing the histograms. AP and AUC between TPs and FPs and the sampling ratio are shown in the upper right corner. 
    (D) NLL histograms of EndoBoost-Finetune. 
  }
  \label{ratio}
\end{figure*}

\begin{table*}[!t]  
  \caption{
    Data efficiency comparison between EndoBoost and ResNet, two post-hocs modules, on the FPPD-13 dataset. In this table, the AP and AUC at eight selected sampling ratios of FPs used during training are presented. The best results of each metric are shown in bold, second-best results are underlined in italics. 
  }
  \centering
  \subtable[AP]{
    \scalebox{0.8}{
    \begin{tabular}{ccccccccc} 
    \toprule  & 1\% & 5\% & 10\% & 20\% & 40\% & 60\% & 80\% & 100\% \\ 
    \midrule \multicolumn{1}{l}{\textbf{ResNet}}  &  &  &  &  &  &  &  &  \\
    Frozen-FC & 0.847$\pm$0.024 & 0.939$\pm$0.011 & 0.946$\pm$0.014 & \textit{\underline{0.959$\pm$0.010}} & 0.963$\pm$0.010 & 0.964$\pm$0.011 & 0.966$\pm$0.011 & 0.968$\pm$0.008 \\
    Frozen-SVM & \textit{\underline{0.887$\pm$0.025\textbf{}}} & \textbf{0.947$\pm$0.007} & \textit{\underline{0.957$\pm$0.001}} & 0.958$\pm$0.009 & \textit{\underline{0.963$\pm$0.006}} & \textit{\underline{0.966$\pm$0.007}} & \textit{\underline{0.967$\pm$0.007}} & \textit{\underline{0.968$\pm$0.007}}  \\
    Finetune & 0.674$\pm$0.061 & 0.774$\pm$0.050 & 0.864$\pm$0.031 & 0.898$\pm$0.024 & 0.899$\pm$0.025 & 0.914$\pm$0.021 & 0.927$\pm$0.006 & 0.933$\pm$0.015 \\
    \multicolumn{1}{l}{\textbf{EndoBoost}} &  &  &  &  &  &  &  &  \\
    Frozen & 0.679$\pm$0.049 & 0.883$\pm$0.037 & 0.894$\pm$0.023 & 0.932$\pm$0.015 & 0.940$\pm$0.016 & 0.955$\pm$0.014 & 0.952$\pm$0.006 & 0.956$\pm$0.007 \\
    Finetune & \textbf{0.897$\pm$0.021} & \textit{\underline{0.943$\pm$0.016}} & \textbf{0.965$\pm$0.010} & \textbf{0.972$\pm$0.010} & \textbf{0.975$\pm$0.006} & \textbf{0.975$\pm$0.009} & \textbf{0.976$\pm$0.007} & \textbf{0.980$\pm$0.007} \\ \bottomrule
    \end{tabular}
    }
  }
  \subtable[AUC]{
    \scalebox{0.8}{
    \begin{tabular}{ccccccccc} 
    \toprule  & 1\% & 5\% & 10\% & 20\% & 40\% & 60\% & 80\% & 100\% \\ 
    \midrule \multicolumn{1}{l}{\textbf{ResNet}}  & ~ & ~ & ~ & ~ & ~ & ~ & ~ & ~ \\
    Frozen-FC & 0.830$\pm$0.030 & 0.943$\pm$0.010 & 0.951$\pm$0.013& 0.963$\pm$0.007 & \textit{\underline{0.967$\pm$0.008}} & \textit{\underline{0.969$\pm$0.007}} & \textit{\underline{0.970$\pm$0.008}} & \textit{\underline{0.972$\pm$0.006}}  \\
    Frozen-SVM & \textit{\underline{0.885$\pm$0.031}} & \textit{\underline{0.946$\pm$0.006\textbf{}}} & \textit{\underline{0.956$\pm$0.003}} & \textit{\underline{0.957$\pm$0.008}} & 0.963$\pm$0.006& 0.966$\pm$0.006 & 0.969$\pm$0.006 & 0.970$\pm$0.005 \\
    Finetune & 0.650$\pm$0.067 & 0.765$\pm$0.045 & 0.854$\pm$0.038 & 0.902$\pm$0.023 & 0.906$\pm$0.024 & 0.915$\pm$0.021 & 0.934$\pm$0.007 & 0.941$\pm$0.014 \\
    \multicolumn{1}{l}{\textbf{EndoBoost}} & ~  & ~ & ~ & ~ & ~ & ~ & ~ & ~ \\
    Frozen & 0.641$\pm$0.087 & 0.876$\pm$0.044 & 0.889$\pm$0.026 & 0.933$\pm$0.020 & 0.941$\pm$0.017 & 0.955$\pm$0.015 & 0.955$\pm$0.007 & 0.957$\pm$0.008  \\
    Finetune & \textbf{0.901$\pm$0.019} & \textbf{0.947$\pm$0.009} & \textbf{0.966$\pm$0.009} & \textbf{0.974$\pm$0.006} & \textbf{0.977$\pm$0.005} & \textbf{0.977$\pm$0.007} & \textbf{0.979$\pm$0.004} & \textbf{0.982$\pm$0.006} \\ \bottomrule
    \end{tabular}
    }
  }
  \label{data_ap_auc}
\end{table*}

Data-efficiency experiments aimed to explore how to make the most use of available FPs during training.  
We compared the model performance between the variants of EndoBoost and ResNet, as shown in Table \ref{data_ap_auc}.
It is clear that the utilization of FPs brought a significant performance boost to the AD approaches trained solely on TPs.
Among the three variants of ResNet utilizing 100\% FPs, ResNet-Frozen-FC and ResNet-Frozen-SVM achieved the highest AP (0.968) and the highest AUC (0.972), respectively.
These classification-based models provide a strong baseline for the quality control task.
In comparison, EndoBoost-Finetune reached the equivalent performance (AP: 0.965, AUC: 0.966) with only 10\% FPs and surpassed it  (AP: 0.972, AUC: 0.974) with 20\% FPs. 
With 100\% FPs used, the EndoBoost-Finetune achieved the highest AP (0.980) and AUC (0.982) among all models.
The performance curves at different sampling ratios (Fig. \ref{ratio}A) demonstrate the superior data efficiency of EndoBoost-Finetune.
As shown in Fig. \ref{ratio}B, the precision-recall curves of EndoBoost-Finetune outperformed all competitors, especially at high sampling ratios.

It is observed that the joint optimization of the feature extractor benefits the performance and data efficiency of EndoBoost.
Compared to EndoBoost-Frozen which used a pre-trained feature extractor, EndoBoost-Finetune achieved higher AP and AUC at all sampling ratios (Fig. \ref{ratio}A).
More intuitively, the likelihood gap between TPs and FPs is widened in EndoBoost-Finetune (Fig. \ref{ratio}C\&D), indicating a more informative feature space for FPs suppression. 
In contrast, fine-tuning the feature extractor harmed FP suppression for ResNet variants. 
As shown in Fig. \ref{ratio}A\&B, ResNet-Finetune was the worst-performing method. 
With a frozen feature extractor, ResNet-Frozen variants were generally better than ResNet-Finetune. 
The performance drop between finetuned and frozen variants was more pronounced when few FPs were used, e.g., an AP decrease of 0.2 when 1\% FPs were used. 
The choice of the SVM of FC did not affect the performance much at high sampling ratios while using SVM brought some advantages over FC when the sampling ratio was low. 

\begin{figure*}[!t]
  \centering
  \includegraphics[width=\textwidth]{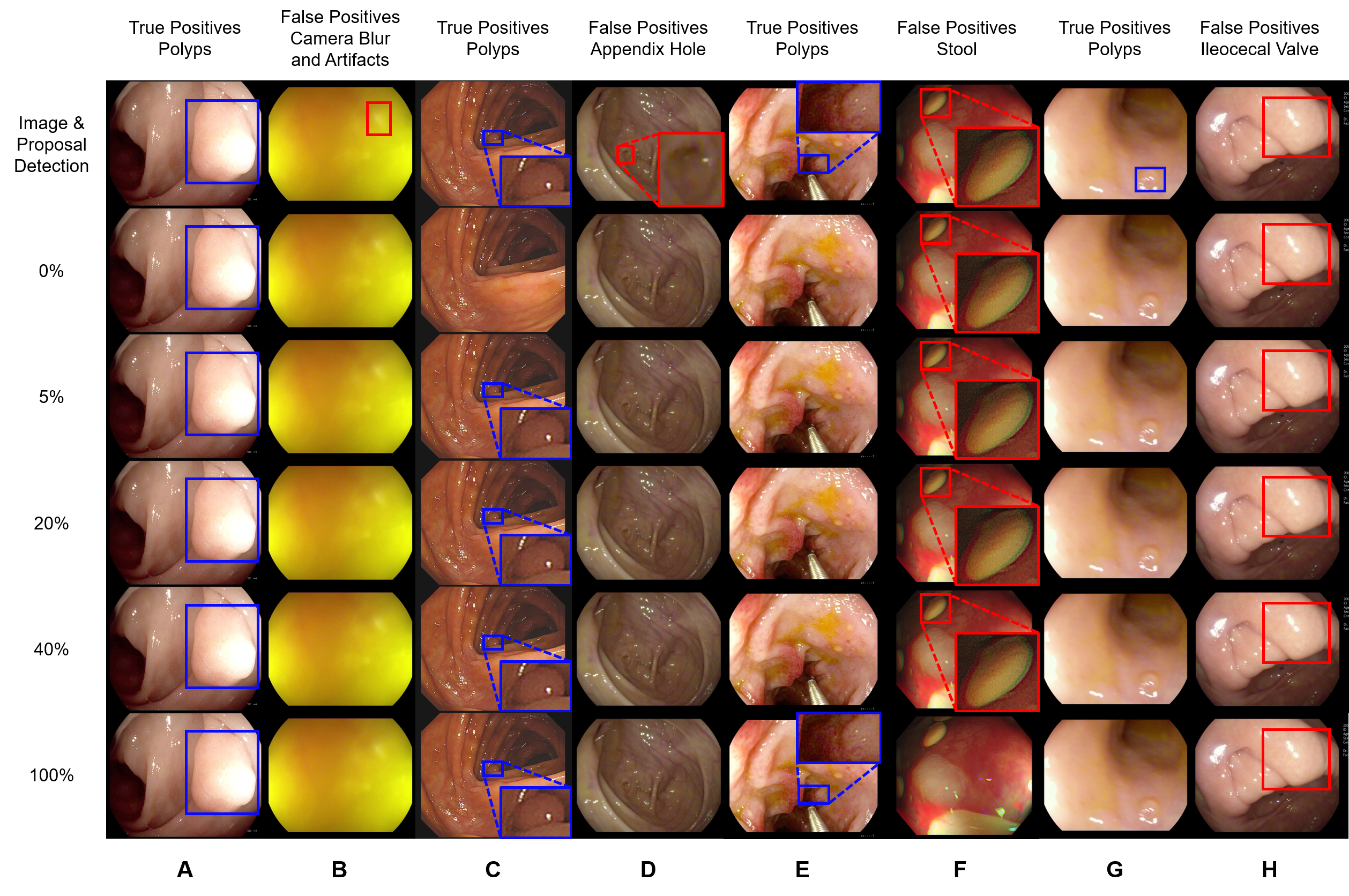}
  \caption{
    Representative samples and results in data-efficiency experiments. The first row shows samples and the proposal detection bounding boxes from the FPPD-13 dataset. The second row is the acceptance/rejection results by EndoBoost-MLE and the third to last row is produced by EndoBoost-Finetune at different sampling ratios. 
    For TP samples with blue bounding boxes, we accept their proposal detections, while the proposal detections should be rejected for the FP samples with red bounding box.  
  }
  \label{ratio_imgs}
\end{figure*}

Fig. \ref{ratio_imgs} illustrates some representative samples of both TPs and FPs in data-efficiency experiments. 
For both TPs and FPs, some easy samples (Col. A-D) could be discriminated with few or even no FPs in training, while hard samples (Col. E-F) could only be correctly accepted/rejected by using an amount of FPs in training. 
However, low-quality TP (Col. G) and FP that is highly similar to TP (Col. H) may still result in failure cases.

\subsection{Robustness to unknown classes of false positives}

\begin{table*}
  \caption{
    Quantitative results for class-robustness experiments. All metrics are means and standard errors of 13 classes of FPs with five-fold cross-validation (a total of 65 experiments). The best results of each metric are shown in bold, second-best results are underlined in italics. 
  }
  \centering
  \scalebox{0.9}{
  \begin{tabular}{ccccccc} 
  \toprule & AP $\uparrow$ & AUC $\uparrow$ & Accuracy $\uparrow$ & Precision $\uparrow$ & Sensitivity $\uparrow$ & Specificity $\uparrow$ \\ 
  \midrule \multicolumn{1}{l}{\textbf{ResNet}}  & & & & & & \\
  Frozen-FC & 0.692$\pm$0.084& \textit{\underline{0.962$\pm$0.014}} & 0.947$\pm$0.010 & 0.635$\pm$0.080 & 0.664$\pm$0.119 & 0.969$\pm$0.006 \\
  Frozen-SVM & \textit{\underline{0.697$\pm$0.126}} & 0.957$\pm$0.026 & \textit{\underline{0.948$\pm$0.019}} & \textit{\underline{0.646$\pm$0.146}} & 0.657$\pm$0.121 & \textit{\underline{0.970$\pm$0.015}}  \\
  Finetune & 0.540$\pm$0.110 & 0.903$\pm$0.043 & 0.931$\pm$0.017 & 0.529$\pm$0.144 & 0.446$\pm$0.106 & 0.968$\pm$0.019 \\
  \multicolumn{1}{l}{\textbf{EndoBoost}} & & & & & &  \\
  MLE & 0.318$\pm$0.180 & 0.793$\pm$0.100 & 0.558$\pm$0.007 & 0.128$\pm$0.013 & \textbf{0.877$\pm$0.092} & 0.534$\pm$0.000 \\
  Frozen & 0.623$\pm$0.116 & 0.932$\pm$0.031 & 0.940$\pm$0.015 & 0.555$\pm$0.163 & 0.572$\pm$0.117 & 0.968$\pm$0.012 \\
  Finetune & \textbf{0.781$\pm$0.080} & \textbf{0.974$\pm$0.011} & \textbf{0.958$\pm$0.011} & \textbf{0.678$\pm$0.110} & \textit{\underline{0.692$\pm$0.097}} & \textbf{0.979$\pm$0.007} \\
  \bottomrule
  \end{tabular}
  }
  \label{class_table}
\end{table*}

\begin{figure*}[!t]
  \centering
  \includegraphics[width=\textwidth]{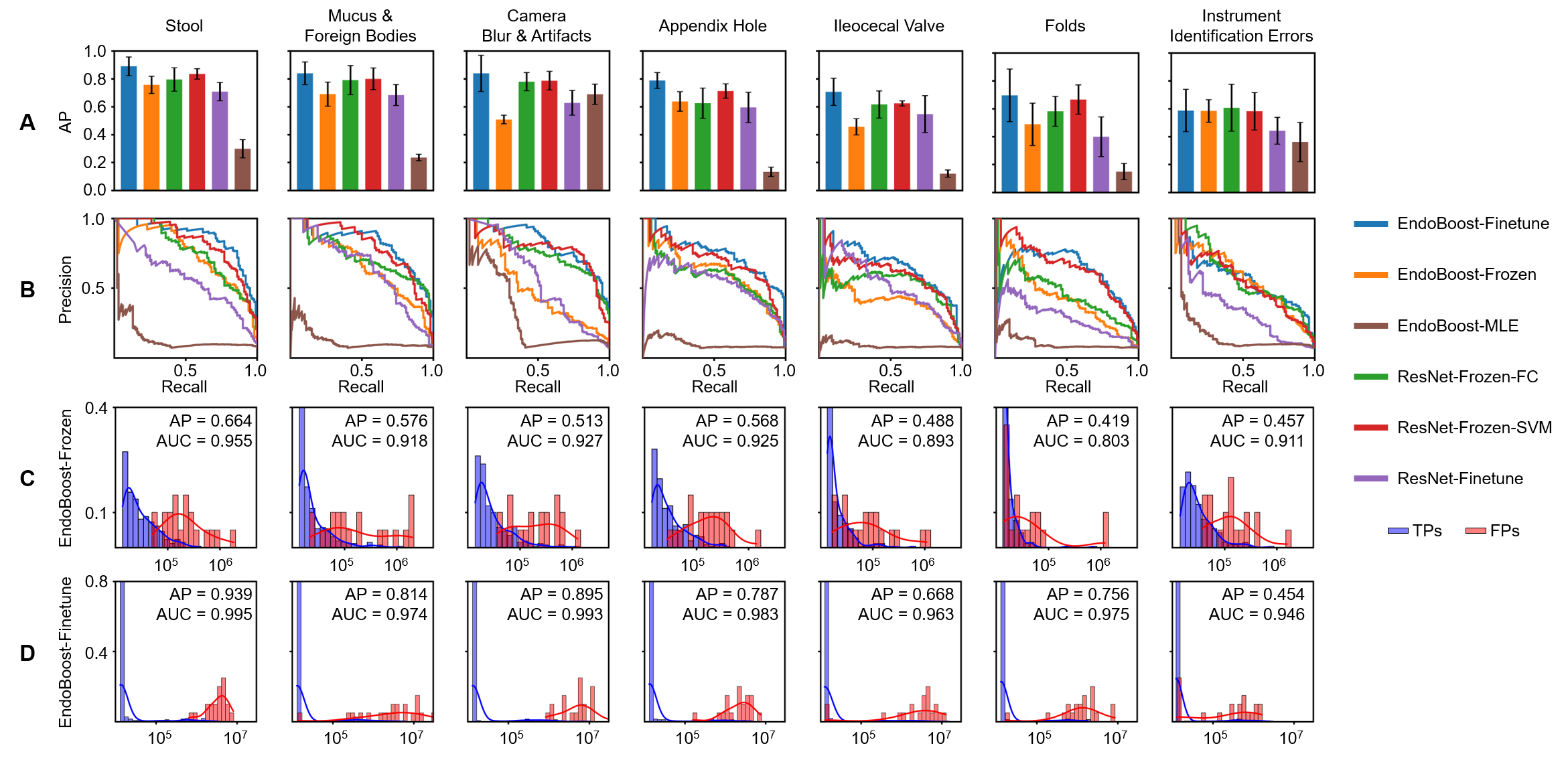}
  \caption{
    Quantitative results for class-robustness experiments. 
    (A) AP for EndoBoost and competitors on selected FP classes. All methods are shown in different colors, the colored bars represent the mean AP for all cross-validation folds while the error bar reflects the standard error. 
    (B) PR curves at different FP classes. Due to the heavy class imbalance, PR curves are zigzagged in class-robustness experiments. 
    (C) NLL histograms for EndoBoost-Frozen.
    (D) NLL histograms for EndoBoost-Finetune. 
  }
  \label{class}
\end{figure*}

The purpose of class-robustness experiments was to evaluate the model robustness to unknown FP categories during training. 
Fig. \ref{class} and Table \ref{class_table} provide quantitative results of the class-robustness experiment. 
Note that TPs and FPs in the class-robustness experiments are imbalanced, there is a gap between the AP and AUC metrics. 
Compared to ResNet classification and other EndoBoost variants, EndoBoost-Finetune achieved the best robustness to unknown FP classes.
As shown in Table \ref{class_table}, EndoBoost-Finetune outperformed other comparison methods with the highest AP (0.781) and AUC (0.974).
Consistent with the observation from data-efficiency experiments, the joint optimization of the feature extractor and normalizing flow also improved the model robustness, which helped EndoBoost achieve higher AP than ResNet classification. 
For a more intuitive illustration, the NLL of TPs and FPs predicted by EndoBoost-Frozen had a large overlap in Fig. \ref{class}C.
Along with the improvement in AP, the NLL overlap between TPs and FPs was significantly reduced with EndoBoost-Finetune.
In contrast, as for different variants of ResNet, finetuning the feature extractor hurt the model robustness, which is also consistent with data-efficiency experiments.
There was also no significant difference between using SVM or FC layer for ResNet classification.

Empirically, the difficulty in distinguishing FPs from different classes can vary widely. 
In Fig. \ref{class}, we present the individual results of seven representative FP classes while the illustration of the remaining six classes can be found in Appendix. 
For all the presented FP classes, EndoBoost-Finetune achieved optimal or sub-optimal performance in the knock-out experiments. 
It is observed that the AP is above 0.8 for the following FPs: stool, mucus \& foreign bodies, and camera blur \& artifact, which can be easily rejected by EndoBoost during real-world colonoscopy.
The FPs from appendix hole, ileocecal valve and folds classes are considered difficult to distinguish from polyps by endoscopists, while the proposed EndoBoost-Finetune achieved satisfactory AP ranging from 0.7 to 0.8.
\begin{figure*}[!t]
  \centering
  \includegraphics[width=\textwidth]{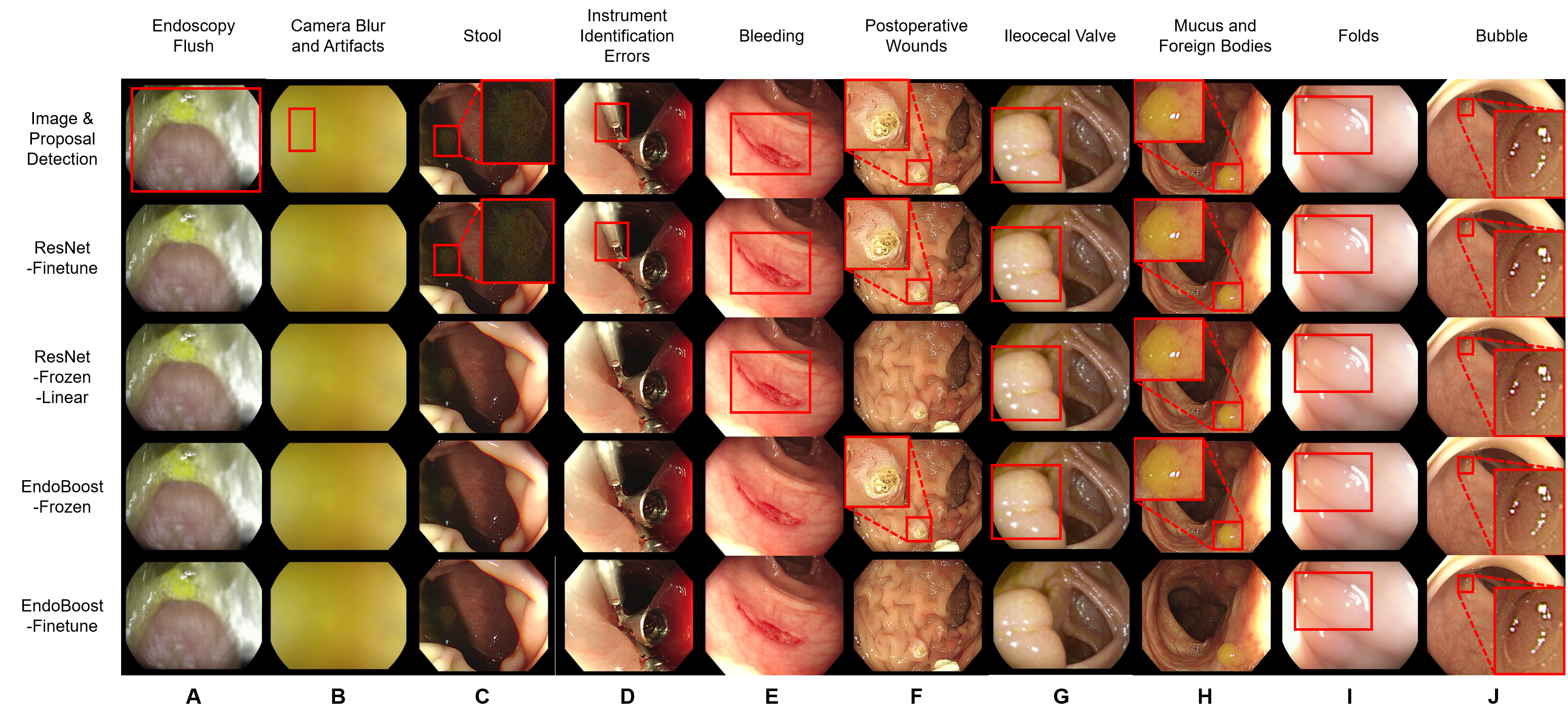}
  \caption{
    Representative FPs and their rejection results in class-robustness experiments. 
    The first row shows samples and the proposal detection bounding boxes from the FPPD-13 dataset, and the second to last row represents rejection results by different methods. 
    All listed samples are FPs and their corresponding proposal detection should be rejected.
    All samples are arranged in difficulty from left to right, the easiest samples are placed on the far left, and vice versa. 
  }
  \label{class_imgs}
\end{figure*}
Representative FPs during class-robustness experiments are shown in Fig. \ref{class_imgs}. Most methods were able to reject the unknown FPs that look significantly different from TPs (Col. A-D). With the unknown FPs more visually alike to the polyps (Col. E-H), some competitors failed to reject them while the EndoBoost-Finetune succeeded. As for hard classes that even confused experts (Col. I\&J), all methods failed to reject such FPs when they are unknown in the training set.

\subsection{Manifold visualization in the feature space}

\begin{figure*}[!t]
  \centering
  \includegraphics[width=\textwidth]{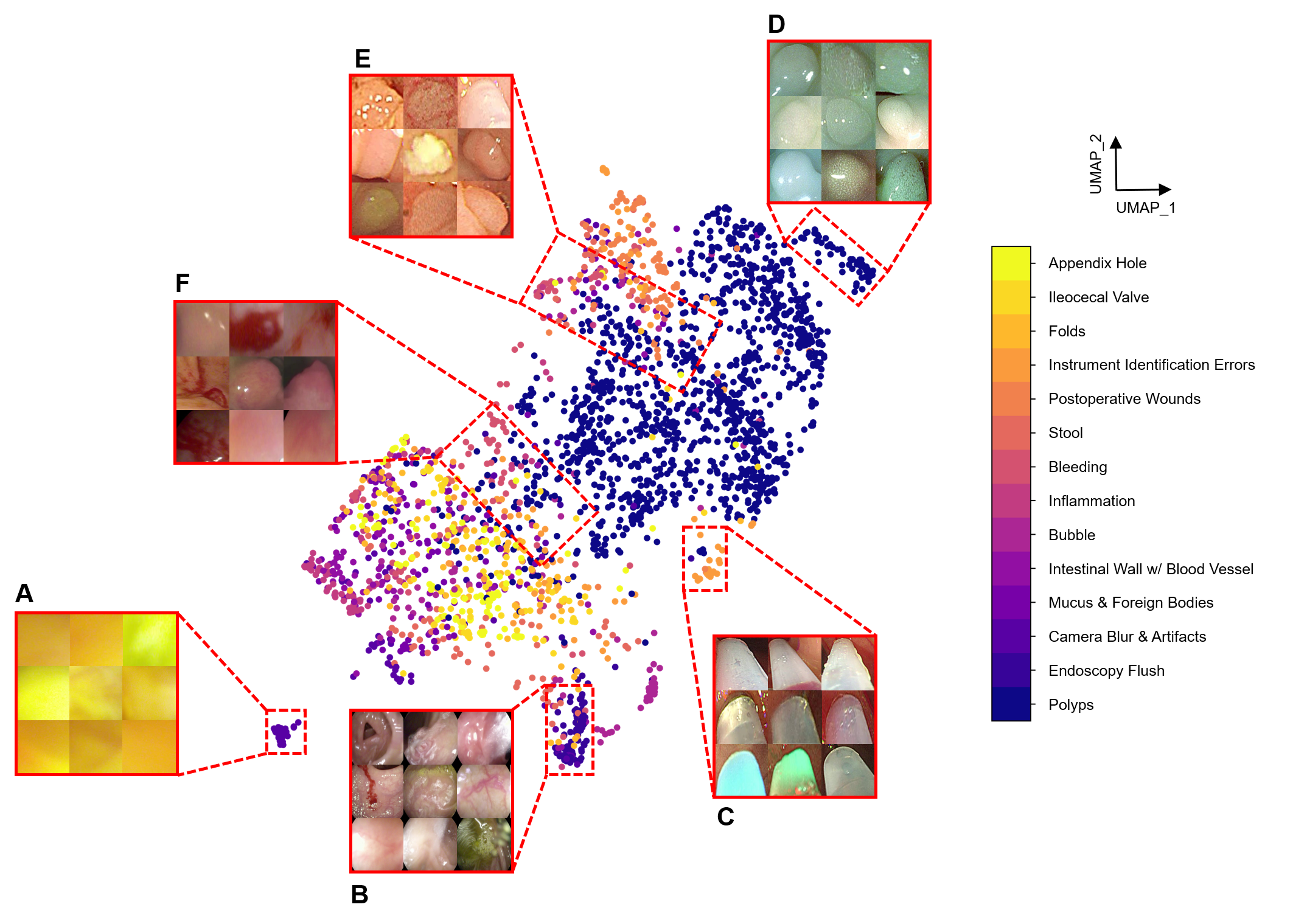}
  \\ \hspace*{\fill} \\
  \caption{
    Two-dimensional UMAP feature visualization of EndoBoost on FPPD-13 dataset. EndoBoost-Finetune trained with all FPs in the training set (sampling ratio of 100\% in data-efficiency experiment) is used for feature extraction, and the color bar on the right represents different FP classes. 
  }
  \label{umap}
\end{figure*}

To better understand the advantage of EndoBoost in distinguishing TPs and FPs, we visualize the samples in the feature space of trained EndoBoost-Finetune. 
The UMAP \citep{mcinnes2018umap} was used for non-linear dimension reduction in Fig. \ref{umap}. 
TPs and FPs are well separated in the feature space, where TPs mostly locate in the upper right corner and the FPs locate in the lower left and upper left corner. 
Different clusters of samples with similar appearances can be observed in the feature space. 
For example, the FPs from camera flare (Fig. \ref{umap}A) are located far from the majority. 
Samples in Fig. \ref{umap}B belong to different FP classes but they share a similar visual appearance. 
The FPs from surgical instruments in Fig. \ref{umap}C are very alike, and the colors of polyps in Fig. \ref{umap}D are green. 
Due to apparent visual distinction, these clusters are far from other samples in the feature space and can be correctly rejected. 
However, for regions where TPs and FPs are intertwined (Fig. \ref{umap} E\&F), the visual similarity of neighboring samples makes the discrimination challenging. 
For example, samples in Fig. \ref{umap}E share a similar orange appearance but they belong to different classes of TPs and FPs. Samples in Fig. \ref{umap}F are generally darker, making it difficult to distinguish polyps from bleeding and inflammation.
In comparison, the samples are also visualized in the feature space of ImageNet pre-trained ResNet in the Appendix where the distributions of FPs and TPs are more mixed.

\begin{figure*}[!t]
  \centering
  \includegraphics[width=\textwidth]{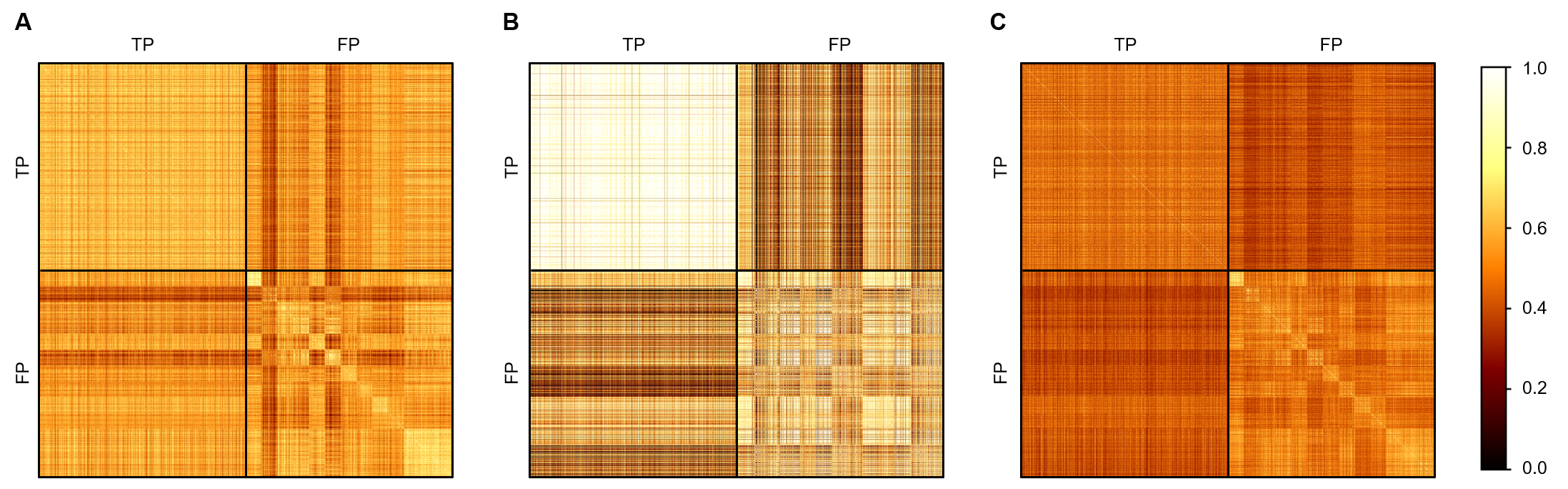}
  \\ \hspace*{\fill} \\
  \caption{
    Cosine similarity matrix of different types of features. To calculate the cosine similarity matrix, we first rank the input features with its label, so the TPs and FPs are placed in the first half and the second half, respectively. The black horizontal and vertical lines indicate the separation of TPs and FPs. In all three matrices, brighter elements correspond to higher similarity, and vice versa. Three types of features are shown: 
    (A) Features extracted with ImageNet pre-trained ResNet.
    (B) ResNet finetuned on FPPD-13. 
    (C) EndoBoost finetuned on FPPD-13. 
  }
  \label{cos_mat}
\end{figure*}

Furthermore, we visualize the cosine similarity matrix of three feature extractors in Fig. \ref{cos_mat}: (A) ImageNet pre-trained ResNet which is the feature extractor for EndoBoost-MLE, EndoBoost-Frozen, and two variants of ResNet-Frozen. (B) ResNet finetuned on FPPD-13 classification, which corresponds to ResNet-Finetune. (C) EndoBoost that finetuned on FPPD-13, which corresponds to EndoBoost-Finetune. 
In Fig. \ref{cos_mat}A, a large number of FPs in the upper right block of the matrix shows high similarity to the TPs in the upper left block, which may result in confusion between TPs and FPs. 
For feature representation of ResNet finetuned on FPPD-13 in Fig. \ref{cos_mat}B, despite the high intra-class similarities of TPs and FP, there are still a certain number of FPs that are similar to TPs in feature space. 
Finally, the feature extractor of EndoBoost-Finetune produces slightly lower intra-class similarity, as shown in the diagonal blocks of Fig. \ref{cos_mat}C, but with a more clear separation between TPs and FPs, resulting in a better performance compared to other methods.

\subsection{Deployment in real-world colonoscopy}
As a post-hoc module for false positive suppression, EndoBoost was further validated in real-world colonoscopy. We took three colonoscopic video clips with a total length of about 1 hour as the real-world deployment test set. 
Representative frames from the YOLOv5 polyp detector with and without EndoBoost false positive reduction are shown in Fig. \ref{video_frames}. 
Video clips can be found in the Supplementary Material.
With the assistance of EndoBoost, FPs were suppressed successfully during the withdrawal of the colonoscopy. 
The effective FP suppression of endoluminal materials (e.g., blood, stool, and bubbles) and artifacts from bowel wall-like tissues (e.g., folds and ileocecal valve) shows the potential of EndoBoost to be integrated into CADe system for clinical use. 

\begin{figure*}[!t]
  \centering
  \includegraphics[width=0.9\textwidth]{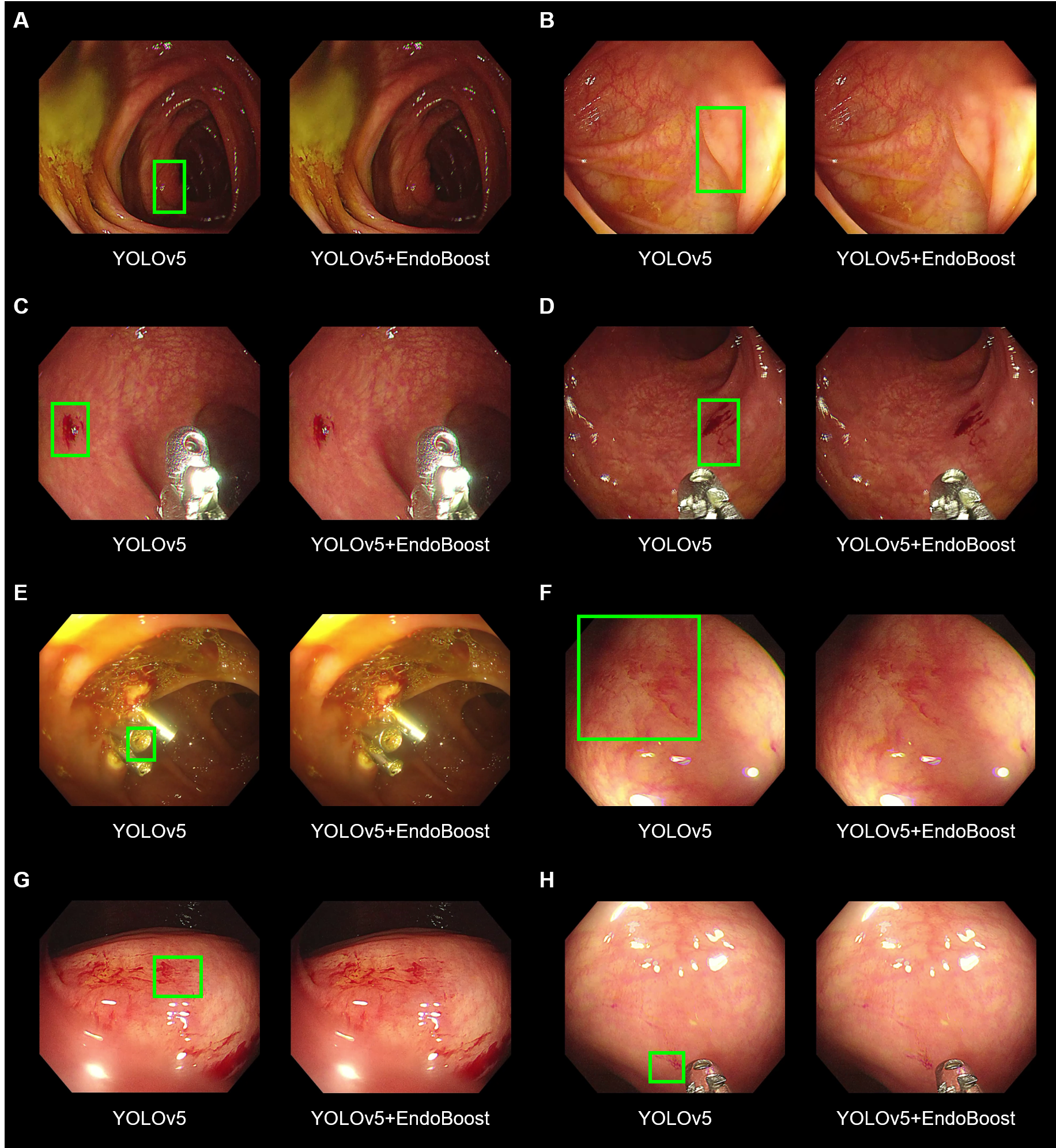}
  \caption{
    Real-world polyp detection by polyp detector alone and appended with EndoBoost. Frame A to H illustrates their behaviors when encountering different types of false positives. 
  }
  \label{video_frames}
\end{figure*}

\section{Discussion}



False positive reduction is a timely need for AI-assisted colonoscopy. 
In this work, we presented solutions from both data and methodology perspectives. 
We introduce the FPPD-13 dataset that contains real-world cases of FPs in colonoscopy and a fine-grained taxonomy of 13 FP classes. 
Furthermore, we propose EndoBoost, a post-hoc module for suppressing false positive predictions during computer-aided polyp detection. 
In comparison with other anomaly detection and classification methods, EndoBoost is better at detecting false positives.
Furthermore, EndoBoost shows promising data efficiency and robustness to unknown false positives.

The curated FPPD-13 dataset provides a collection of false positive predictions produced by a SOTA polyp detector in real-world colonoscopy, which is distinct from previous public colonoscopic datasets. 
FPPD-13 is a versatile dataset for endoscopic image analysis. 
First of all, FPPD-13 can serve as a benchmark dataset to evaluate the robustness of different AI-assisted polyp detectors in face of real-world artifacts. 
Besides, since FPPD-13 provides 2,600 images and half of them are false positives, it could be a valuable addition to the current dataset when training or fine-tuning the polyp detector. 
Furthermore, as we demonstrated in this work, FPPD-13 could help develop new post-hoc modules for false positive suppression. 

The superior performance of EndoBoost comes from its ability to model complex high-dimensional probability distributions. 
The likelihood, a quantitative measure of density in the language of probability, indicates how the samples are distributed in the feature space. 
Since TPs and FPs follow different distributions in the feature space as shown in Fig. \ref{umap}, the likelihood is a convenient indicator for false positive suppression, while the reconstruction error or distance to the decision boundary is prone to bias in the anomaly detection task. 
In comparative experiments, EndoBoost-MLE and KDE were both density-based methods, however, KDE suffers from the curse of dimensionality, which explains the large deterioration between EndoBoost-MLE and KDE in Table \ref{comp_tab}. 
The classification-based methods were sub-optimal among all the AD methods. 
OC-SVM and MCD could learn proper decision boundaries due to the obvious difference in features between TPs and FPs. 
For the reconstruction-based methods, the performance of AE was generally worse than PCA. An explanation is that AE has a better ability in reconstruction, which caused the reconstruction errors of both TPs and FPs to be low, while the linear nature of PCA enlarged the difference in reconstruction error between TPs and FPs. 
However, VAE was the best among all reconstruction-based methods, indicating the importance of appropriate regularization. 

Compared with ResNet classifiers, EndoBoost demonstrates superior performance and data efficiency in utilizing FPs. 
EndoBoost with outlier exposure could also be considered as a binary classifier.
The explicit modeling of data distribution not only benefits anomaly detection but also can be used as a regularization term to prevent overfitting. 
In contrast, the ResNet only learned a decision boundary between TPs and FPs but lacked the exploration of data distribution. 
This explains why EndoBoost-Finetune outperforms EndoBoost-Frozen while ResNet-Finetune is inferior to ResNet-Frozen-FC and ResNet-Frozen-SVM. 
Since the density estimation could be considered as a regularization task, finetuning the feature extractor did not result in overfitting. 
Instead, the joint optimization helped obtain an informative feature space suitable for the task of FPs suppression. 
EndoBoost is also more robust to unknown classes of FPs than its binary classification counterpart. 
Binary classifiers using ResNet divide the high-dimensional space into two halves with the decision boundary, which may easily misclassify the unknown classes of FPs. 
However, the density-aware EndoBoost is naturally robust to unknown FP classes since it learns the structure of TPs distribution. 


The precise estimation of likelihood itself is also a sign of better interpretability. 
Samples with likelihoods far from the threshold are considered to be easy samples, while samples with likelihoods that is close to the threshold may require the endoscopists' involvement for a better decision.
During real-world deployment, the choice of likelihood threshold is of vital importance.
Likelihood thresholds that are too low might fail to suppress FPs since it accepts the most positive predictions. 
Excessive thresholds reject more FP predictions, but also cause missing detection of polyps. 
An ideal threshold should be high enough to filter out common FPs but hardly reject TPs, preserving the sensitivity of the original polyp detector.
From the PR curve on the test set, we found that EndoBoost-Finetune could filter out 49\% FPs safely without rejecting any TP.
In fact, an advantage of the EndoBoost is its simplicity and flexibility in setting a threshold according to the clinical requirement.


Although EndoBoost achieved a satisfactory performance of FP suppression in extensive experiments, whether it can really improve the endoscopic procedure remains real-world validation. 
Further clinical trials on the adenoma detection rate, withdrawal time, and satisfaction degree of endoscopists are being carried out.

\section{Conclusion}

In this work, we present a practical solution of dataset and methodology for reducing FPs during AI-assisted colonoscopy. 
We introduce the FPPD-13 dataset which contains 13 classes of FPs during real-world polyp detection. 
We also propose EndoBoost, a plug-and-play module to filter out FPs with density estimation in an informative feature space. 
It exceeds the performance of fully supervised classifiers using only 20\% of FPs and is more robust to unknown FP classes. 
For future work, we plan to extend the input of EndoBoost to the video stream and further combine spatiotemporal information for better quality control of object detection. 
Besides, multi-center clinical trials using EndoBoost are being carried out.

\section*{Acknowledgments}
We thank Te Luo, Wu-Chao Tao, Wen-Long Wu, Zi-wei Li, De-Jia Sun, and Jia-Yan Wang for their kindness and support of this research.
This study was supported by grants from the National Natural Science Foundation of China (82203193) and the Shanghai Sailing Programs of Shanghai Municipal Science and Technology Committee (22YF1409300).

\bibliographystyle{model2-names.bst}\biboptions{authoryear}
\bibliography{arxiv.bbl}

\newpage

\begin{appendices}
\label{appendix_A}
\renewcommand{\thetable}{\Alph{section}\arabic{table}}
\renewcommand{\thefigure}{\Alph{section}\arabic{figure}}
\setcounter{table}{0}
\setcounter{figure}{0}

\section{Performance validation of YOLOv5 polyp detector}

We evaluated the well-trained YOLOv5 polyp detector on a widely used challenge dataset CVC-ClinicDB \citep{bernal2017comparative}.
Following the evaluation metrics in \cite{wang2018development}, we report the number of true positives, false negatives, true negatives, false positives, sensitivity, and specificity in Table \ref{yolov5_public_private}. 
Our polyp detector shows excellent ability in detecting polyps in the public dataset.

\begin{table*}[!tb]
\caption{Performance of YOLOv5 polyp detector on a public colonoscopy datasets.}
\centering
\scalebox{0.9}{
\begin{tabular}{ccccccccc}
\toprule
Dataset & 
Method & 
\makecell[c]{Total number of\\ images} & 
\makecell[c]{True\\ positives} & 
\makecell[c]{False\\ negatives} & 
\makecell[c]{True\\ negatives} & 
\makecell[c]{False\\ positives} & 
\makecell[c]{Sensitivity} \\ 
\midrule
\multirow{3}*{\makecell[c]{CVC-ClinicDB \\ \citep{bernal2017comparative}}} & 
\cite{wang2018development} & \multirow{3}*{612} & 570 & 76 & NA & 42 & 88.24\% \\
 & \cite{lee2020real} &  & 577 & 63 & NA & 10 & 90.16\% \\
 & Ours &  & 626 & 20 & NA & 40 & 96.90\% \\
\bottomrule
\end{tabular}}
\label{yolov5_public_private}
\end{table*}

\section{Supplementary figures}

\begin{figure*}[!t]
  \centering
  \includegraphics[width=0.9\textwidth]{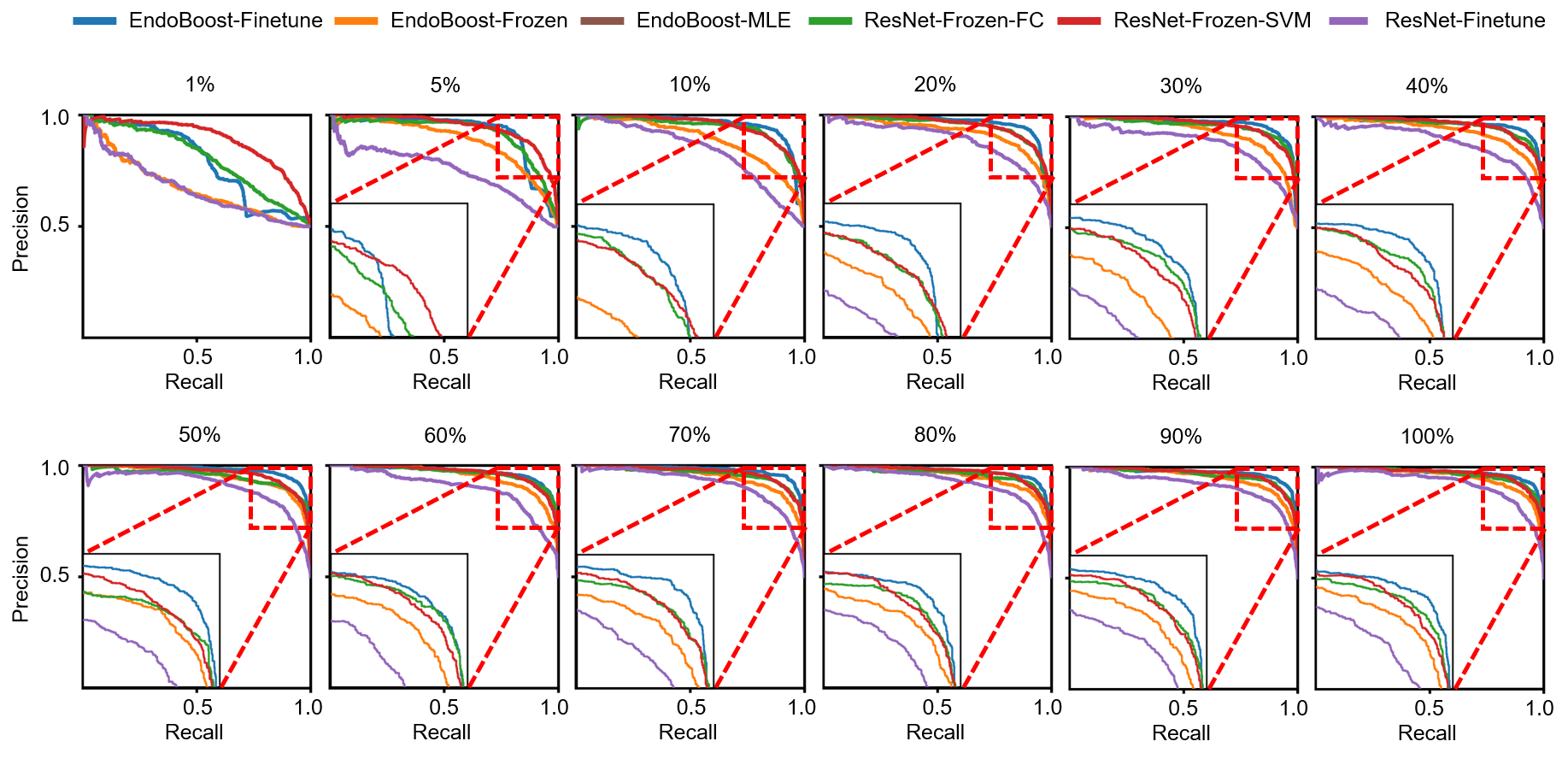}
  \caption{
    PR curves in data-efficiency experiments for all 12 sampling ratios.
  }
  \label{supp_data_pr}
\end{figure*}

\begin{figure*}[!t]
  \centering
  \includegraphics[width=0.9\textwidth]{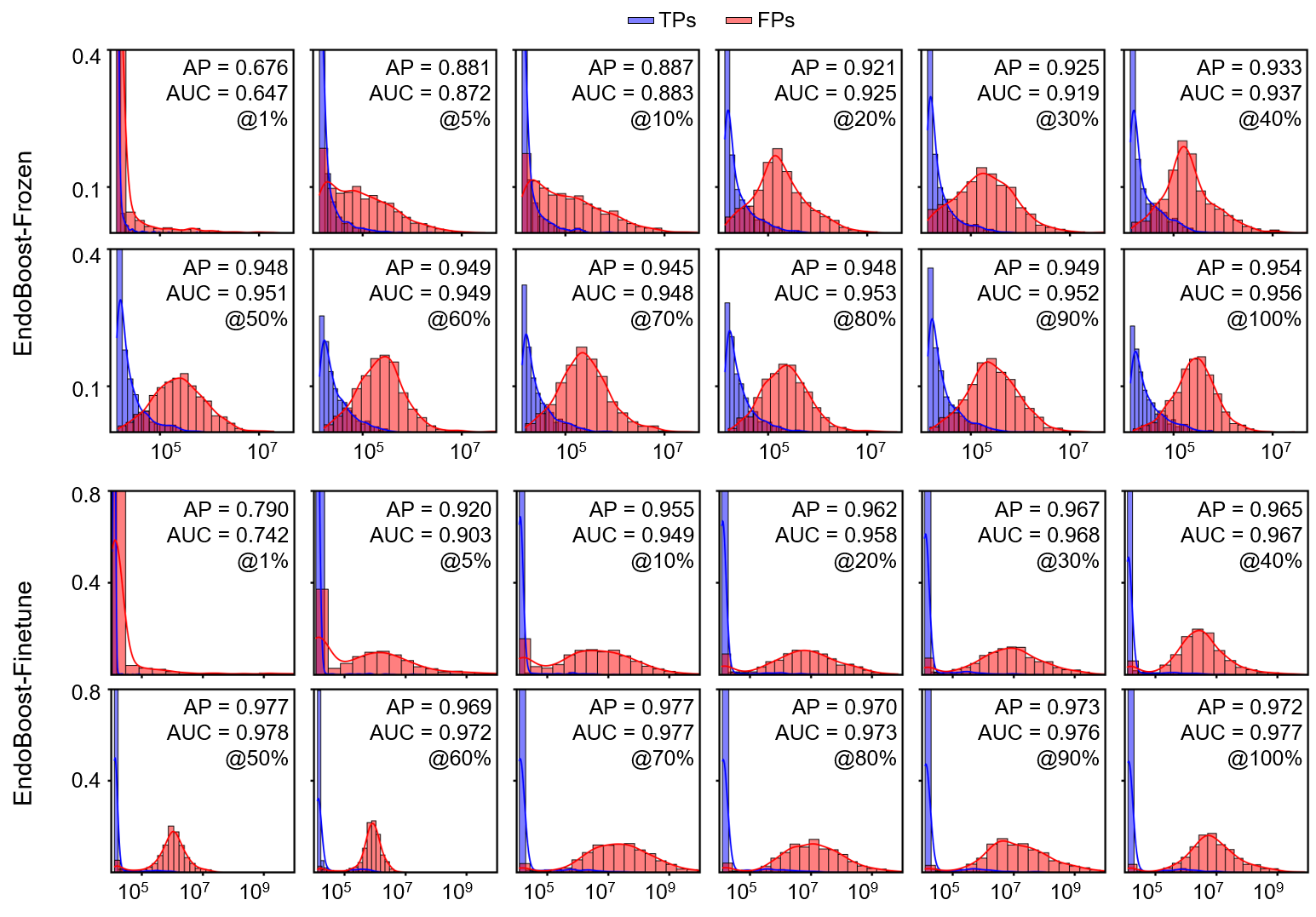}
  \caption{
    NLL histograms in data-efficiency experiments for EndoBoost-Frozen and EndoBoost-Finetune.
  }
  \label{supp_data_nlls}
\end{figure*}

\begin{figure*}[!t]
  \centering
  \includegraphics[width=\textwidth]{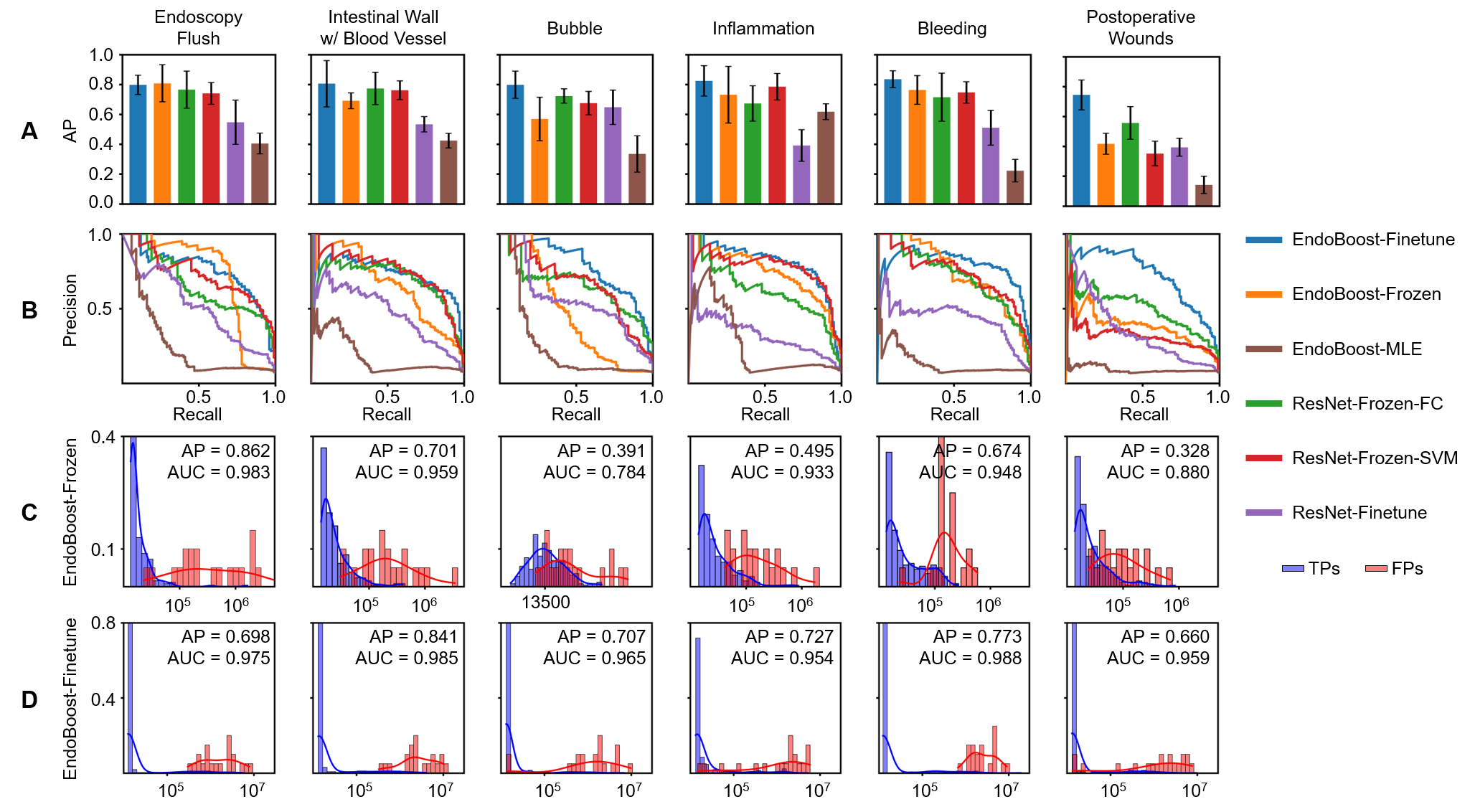}
  \caption{
    Quantitative results in class-robustness experiments for the other six remaining false positive classes.
  }
  \label{supp_class_exps}
\end{figure*}

\begin{figure*}[!t]
  \centering
  \includegraphics[width=0.9\textwidth]{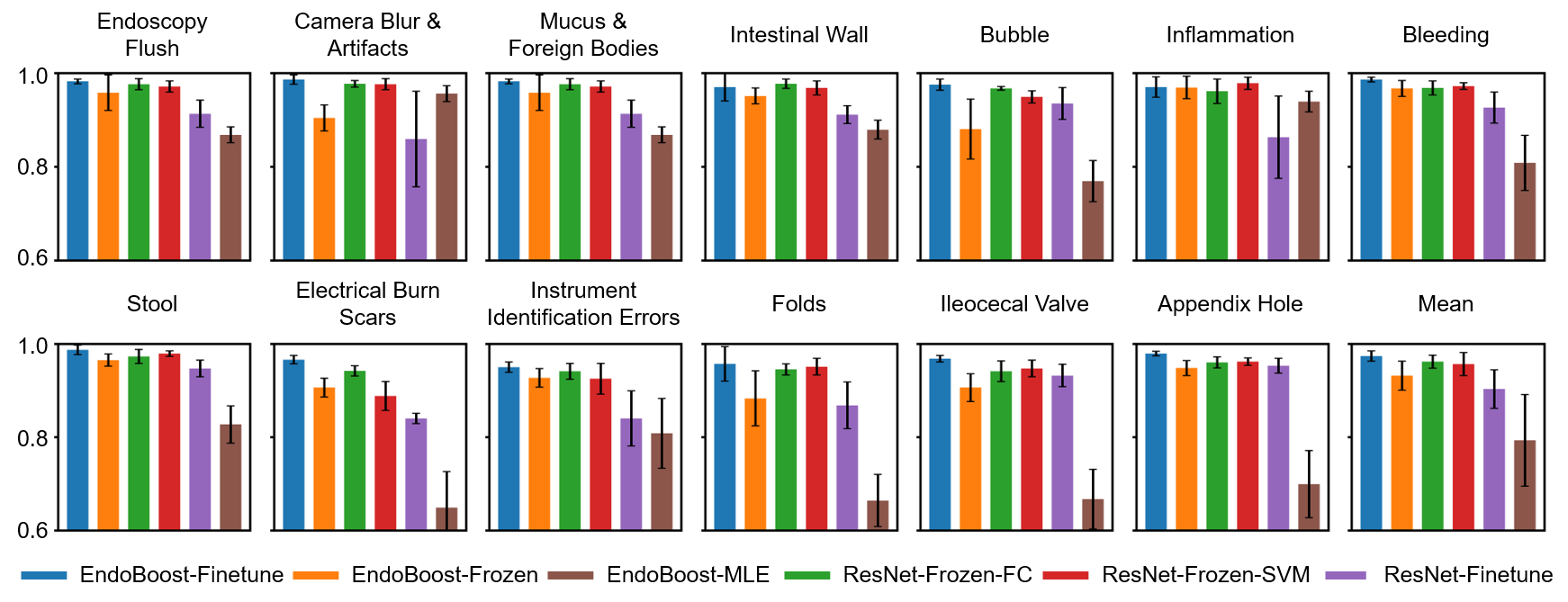}
  \caption{
    AUC in class-robustness experiments for all FP classes. The average AUC for all FP classes is also shown.
  }
  \label{supp_class_auc}
\end{figure*}

\begin{figure*}[!t]
  \centering
  \includegraphics[width=\textwidth]{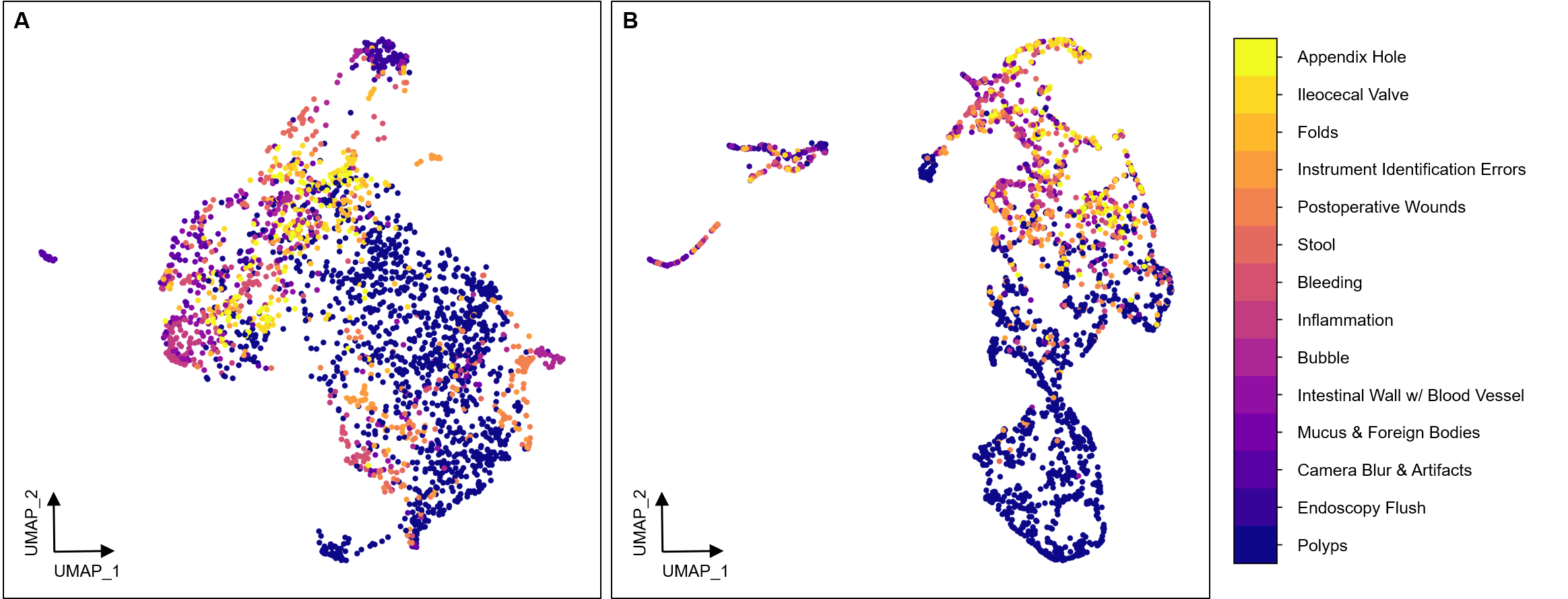}
  \caption{
    2D UMAP feature visualization on FPPD-13 dataset. (A) Feature of ImageNet pre-trained ResNet, which is used in EndoBoost-MLE, EndoBoost-Frozen, ResNet-Frozen-SVM, ResNet-Frozen-Linear, and comparative AD methods. (B) Feature of ResNet-Finetune on FPPD-13.
  }
  \label{supp_umap}
\end{figure*}

We present some supplementary figures to better illustrate the experimental results. 
In Fig. \ref{supp_data_pr}, PR curves in data-efficiency experiments at all 12 sampling ratios are shown.
Despite the inferiority when little FPs are used in training, EndoBoost-Finetune has been the best method since 10\% sampling ratio.
In Fig. \ref{supp_data_nlls}, EndoBoost-Frozen and EndoBoost-Finetune quickly converge to high AP of FP suppression in data-efficiency experiments.
What's more, EndoBoost-Finetune reaches higher performance than EndoBoost-Frozen with only 20\% FPs used.
For class-robustness experiments in Fig. \ref{supp_class_exps}, EndoBoost-Finetune is the most robust method in almost all FP classes and the joint optimization significantly improve the robustness of EndoBoost.
AUC of individual FP classes and the mean AUC are also provided in Fig. \ref{supp_class_auc}, which shows a consistent result with AP.  
In Fig. \ref{supp_umap}, 2D UMAP feature visualizations of ResNet50 with ImageNet pre-trained weight (Fig. \ref{supp_umap}A) and ResNet50 finetuned on FPPD-13 (Fig. \ref{supp_umap}B) are also shown.

\end{appendices}

\end{document}